\documentclass[lettersize,journal]{IEEEtran}
\usepackage{amsmath,amsfonts}
\usepackage{algorithmic}
\usepackage{algorithm}
\usepackage{array}
\usepackage[caption=false,font=normalsize,labelfont=sf,textfont=sf]{subfig}
\usepackage{textcomp}
\usepackage{stfloats}
\usepackage{url}
\usepackage{verbatim}
\usepackage{graphicx}
\usepackage{booktabs}
\usepackage{cite}

\usepackage{multirow}
\usepackage[compact]{titlesec}
\usepackage{xcolor}
\usepackage{amsthm}
\newtheorem{theorem}{Theorem}
\newtheorem{corollary}{Corollary}

\usepackage{threeparttable}
\usepackage{tcolorbox}
\usepackage{fancyvrb}
\usepackage{fvextra}
\begin{document}

\title{LLM-Driven Stationarity-Aware Expert Demonstrations for Multi-Agent Reinforcement Learning in Mobile Systems}

\author{
    Tianyang Duan,
    Zongyuan Zhang,
    Zheng Lin,
    Songxiao Guo,
    Xiuxian Guan,
    Guangyu Wu,
    Zihan Fang,
    Haotian Meng,
    Xia Du, Ji-Zhe Zhou,
    Heming Cui,~\IEEEmembership{Member,~IEEE},
    Jun Luo,~\IEEEmembership{Fellow,~IEEE},
    and 
    Yue Gao,~\IEEEmembership{Fellow,~IEEE}
    \thanks{T. Duan, Z. Zhang, S. Guo, X. Guan, and H. Cui are with the Division of Computer Science, The University of Hong Kong, Hong Kong SAR, China (e-mail: tyduan@cs.hku.hk; zyzhang2@cs.hku.hk; sxguo@connect.hku.hk; xxguan@cs.hku.hk; heming@cs.hku.hk).}
    \thanks{Z. Lin is with the Department of Electrical and Electronic Engineering, The University of Hong Kong, Hong Kong SAR, China (e-mail: linzheng@eee.hku.hk).}
    \thanks{G. Wu is with the Department of Computer Science and Technology, Peking University, Beijing, China (e-mail: gywu9908@163.com).}   
    \thanks{Z. Fang is with the Department of Computer Science, City University of Hong Kong, Hong Kong SAR, China (e-mail: zihanfang3-c@my.cityu.edu.hk).}
    \thanks{H. Meng is an employee in the China Unicom Digital Technology, China Unicom co.,Ltd, China (e-mail: 790498573@qq.com).}
     \thanks{ X. Du is with the School of
 Computer and Information Engineering, Xiamen University of Technology, Xiamen, China (email: duxia@xmut.edu.cn).}
      \thanks{ J. Zhou is with the School of Computer Science, Engineering Research Center of Machine Learning and Industry Intelligence, Sichuan University, Chengdu, China (email: yb87409@um.edu.mo).}   
    \thanks{J. Luo is with the College of Computing and Data Science, Nanyang Technological University, Singapore (e-mail: junluo@ntu.edu.sg).}
    \thanks{Y. Gao is with the Institute of Space
 Internet, Fudan University, Shanghai, China, and the School of
 Computer Science, Fudan University, Shanghai, China (e-mail: gao.yue@fudan.edu.cn).} 
}

\maketitle

\begin{abstract}
Multi-agent reinforcement learning (MARL) has been increasingly adopted in many real-world applications. While MARL enables decentralized deployment on resource-constrained edge devices, it suffers from severe non-stationarity due to the synchronous updates of agent policies. This non-stationarity results in unstable training and poor policy convergence, especially as the number of agents increases. In this paper, we propose RELED, a scalable MARL framework that integrates large language model (LLM)-driven expert demonstrations with autonomous agent exploration. RELED incorporates a Stationarity-Aware Expert Demonstration module, which leverages theoretical non-stationarity bounds to enhance the quality of LLM-generated expert trajectories, thus providing high-reward and training-stable samples for each agent. Moreover, a Hybrid Expert-Agent Policy Optimization module adaptively balances each agent's learning from both expert-generated and agent-generated trajectories, accelerating policy convergence and improving generalization. Extensive experiments with real city networks based on OpenStreetMap demonstrate that RELED achieves superior performance compared to state-of-the-art MARL methods.
\end{abstract}


\section{Introduction}

With the significant enhancement of computational capabilities in edge devices such as vehicles and robots, their ability to execute complex tasks and achieve real-time interactions has been substantially improved~\cite{khan2020edge,duan2025sample,lin2024fedsn,yuan2024satsense,tang2024merit,peng2024sums,fang2025dynamic,yuan2025constructing,lin2024adaptsfl,yuan2023graph,lin2025leo}. Multi-agent systems (MAS) have attracted increasing attention due to their unique advantages in distributed collaboration and scalability. Among various MAS paradigms, multi-agent reinforcement learning (MARL) has emerged as a prominent direction in recent years~\cite{ning2024survey,duan2025leed}. MARL leverages lightweight neural networks as policy functions for each agent, markedly reducing inference overhead and latency, and thus facilitating deployment on resource-constrained edge devices~\cite{chen2024communication}. Furthermore, MARL typically employs centralized value estimation for joint policy optimization during training, while enabling decentralized decision-making based on local observations during inference~\cite{xiao2022asynchronous}. This paradigm effectively achieves low-latency responses and minimizes inter-agent communication overhead, leading to its successful application in domains such as computation offloading~\cite{gao2022large, lin2025hierarchical,wei2025optimizing,lin2024split}, drone scheduling~\cite{du2025multi, gao2024moipc}, and urban traffic management~\cite{zhang2023learning,wang2020stmarl}.

Despite these advantages, MARL still faces significant non-stationarity challenges~\cite{zhang2021multi}. Existing MARL training frameworks can be broadly categorized into fully decentralized frameworks~\cite{yu2022surprising,de2020independent} and centralized training with decentralized execution (CTDE)~\cite{rashid2020monotonic,ackermann2019reducing}. In fully decentralized settings, each agent optimizes its policy from local observations and individual rewards. This independence ignores cross-agent behavioral dependencies and results in policy conflicts that degrade overall system performance. CTDE mitigates these issues by using centralized value estimation during training to promote cooperative behavior. Nevertheless, as the scale of the multi-agent system increases, the joint state-action space grows exponentially, leading to a dramatic increase in the cost of policy evaluation and exacerbating estimation errors of each agent’s action value~\cite{wang2020breaking}. The resulting bias often drives agents toward homogenized policies, reducing behavioral diversity and hindering exploration of globally optimal solutions. Fundamentally, these issues stem from learning-induced non-stationarity. Each agent’s environment includes other agents that are simultaneously updating their policies. Any update by one agent alters the environment dynamics observed by the others, violating the stationarity assumptions underlying standard reinforcement learning algorithms~\cite{hernandez2017survey}.

Fortunately, large language models (LLMs) trained on extensive corpora have shown strong reasoning and decision-making in complex settings~\cite{pan2024agentcoord,lin2023pushing,wang2023describe,lin2025hsplitlora,fang2024automated,lin2024splitlora}, making them promising tools for generating expert knowledge and demonstrations to improve policy optimization in MARL. However, LLMs operate from prompts and prior knowledge rather than direct interaction with the environment, they offer limited exploration and thus tend to recommend actions only for familiar or well-specified states~\cite{ahn2022can}. This reduces demonstration diversity and state-space coverage, which in turn hampers policy generalization to unvisited or poorly observed states in MARL. Moreover, the quality of LLM-generated demonstrations is highly dependent on the input context~\cite{liu2022makes}. In partially observable environments, where agents can only obtain incomplete observations of the true state, LLM-generated demonstrations tend to guide agents toward suboptimal behaviors, impeding convergence to optimal policies.

To address these research gaps, we propose RELED, a scalable multi-agent policy optimization framework that integrates LLM-driven expert demonstrations with agent autonomy. Firstly, we develop a Stationarity-Aware Expert Demonstration (SED) module in RELED based on LLMs, which quantifies and theoretically bounds the environmental non-stationarity induced by simultaneous multi-agent policy updates. Refined by feedback metrics derived from our theoretical analysis, the LLM in SED module generates targeted demonstration results for each agent, aiming to maximize high cumulative rewards while constraining non-stationarity. Secondly, we develop a Hybrid Expert-Agent Policy Optimization (HPO) module in RELED. In HPO module, each agent combines LLM-generated demonstrations from the SED module with its own autonomous exploration experiences by independently optimizing a hybrid policy loss function. This hybrid policy optimization approach not only accelerates policy convergence but also mitigates the sample diversity and performance limitations in purely LLM-generated demonstrations, thus improving both learning efficiency and generalization for each agent. The key contributions of this work are summarized as follows:

\begin{itemize}
\item We propose RELED, a scalable MARL framework that couples LLM-based expert demonstrations with decentralized exploration, achieving efficient training.
\item We derive a performance bound with two computable metrics, namely the reward volatility index and the policy divergence index, to quantify non-stationarity.
\item We propose a stationarity-aware LLM refinement mechanism, which uses the above metrics as feedback to iteratively update instructions to generate high-quality demonstrations.
\item We propose a hybrid policy optimization mechanism that transitions from imitation to exploration by adaptively balancing the weights of expert and agent-generated samples.
\item We empirically evaluate RELED on real city networks based on OpenStreetMap. The results demostrate that RELED outperforms state-of-the-art MARL baselines.
\end{itemize}

The rest of the paper is organized as follows. Section~\ref{Sec: Related Work} discusses related work. Section~\ref{Sec: RELED Framework} presents the RELED framework. Section~\ref{Sec: Implementation} describes the implementation of RELED. Section~\ref{Sec: Evaluation} reports the evaluation results. Finally, Section~\ref{Sec: Conclusion} concludes the paper.

\section{Related Work}
\label{Sec: Related Work}
MARL has attracted significant attention for its ability to address complex cooperative and competitive tasks involving multiple agents \cite{zhang2021multi}. A prominent category of MARL algorithms is value-based methods, which estimate the expected returns of joint actions through action-value functions (Q-functions) and derive optimal joint policies accordingly \cite{qu2019exploiting,yuan2023multi,sunehag2017value}. QMIX enhances decentralized policy learning in MARL by employing mixing networks to combine individual agent Q-values into a joint action-value function while maintaining monotonicity \cite{rashid2020monotonic}. Building on QMIX, HMDQN integrates hierarchical reinforcement learning to mitigate sparse reward challenges in multi-robot tasks \cite{bai2023smart}. Another key category is policy-based methods, which directly optimize stochastic policies represented as probability distributions over actions conditioned on states \cite{yu2022surprising,iqbal2019actor}. HATRPO extends trust region policy optimization to multi-agent scenarios, ensuring stable and efficient collaborative learning through advantage decomposition and sequential policy updates \cite{kuba2021trust}. MACPO incorporates safety constraints into MARL, ensuring agents consistently satisfy constraints during policy updates via trust region restrictions \cite{gu2021multi}. Other methods employing deterministic policies output actions directly rather than probability distributions, improving policy robustness by leveraging global information for centralized training \cite{lowe2017multi,ackermann2019reducing}.

Recent advances in LLMs have broadened their knowledge and reasoning abilities, leading to research on using LLMs to improve the performance of single-agent reinforcement learning \cite{pternea2024rl}. Some studies focus on achieving effective representation mapping between LLMs' text-based input-output and the RL environment's state and action spaces \cite{feng2023llama,wu2023spring}. Qiu et al. \cite{qiu2023embodied} propose a multimodal vision-text framework for aligning non-text observations with actions via joint fine-tuning. LANCAR addresses ambiguous robotic locomotion instructions by leveraging LLMs to generate context-aware embeddings \cite{shek2024lancar}. LLaRP uses frozen pre-trained LLMs with learnable adapters to enable strong generalization in vision-language policies \cite{szot2023large}. Other studies have explored LLMs for designing reward functions, addressing the limitations of traditional hand-crafted rewards in terms of sparsity, bias, and scalability \cite{sontakke2023roboclip,triantafyllidis2024intrinsic}. Kwon et al. proposed using LLMs as proxy reward functions, allowing users to define RL objectives with natural language prompts and few-shot examples, thus reducing the need for expert demonstrations \cite{kwon2023reward}. Song et al. developed a self-optimizing reward framework where LLMs iteratively generate and refine reward functions from natural language instructions, replacing manual reward design in continuous control tasks \cite{song2023self}. Several studies \cite{zhu2025lamarl,liu2025llm} have leveraged LLMs to enhance cooperation or improve sample efficiency in MARL. However, leveraging LLMs to address the non-stationarity bottleneck in MARL systems has received little attention and remains an unexplored direction.

\begin{figure*}[ht]
    \centering
    \includegraphics[width=0.85\linewidth]{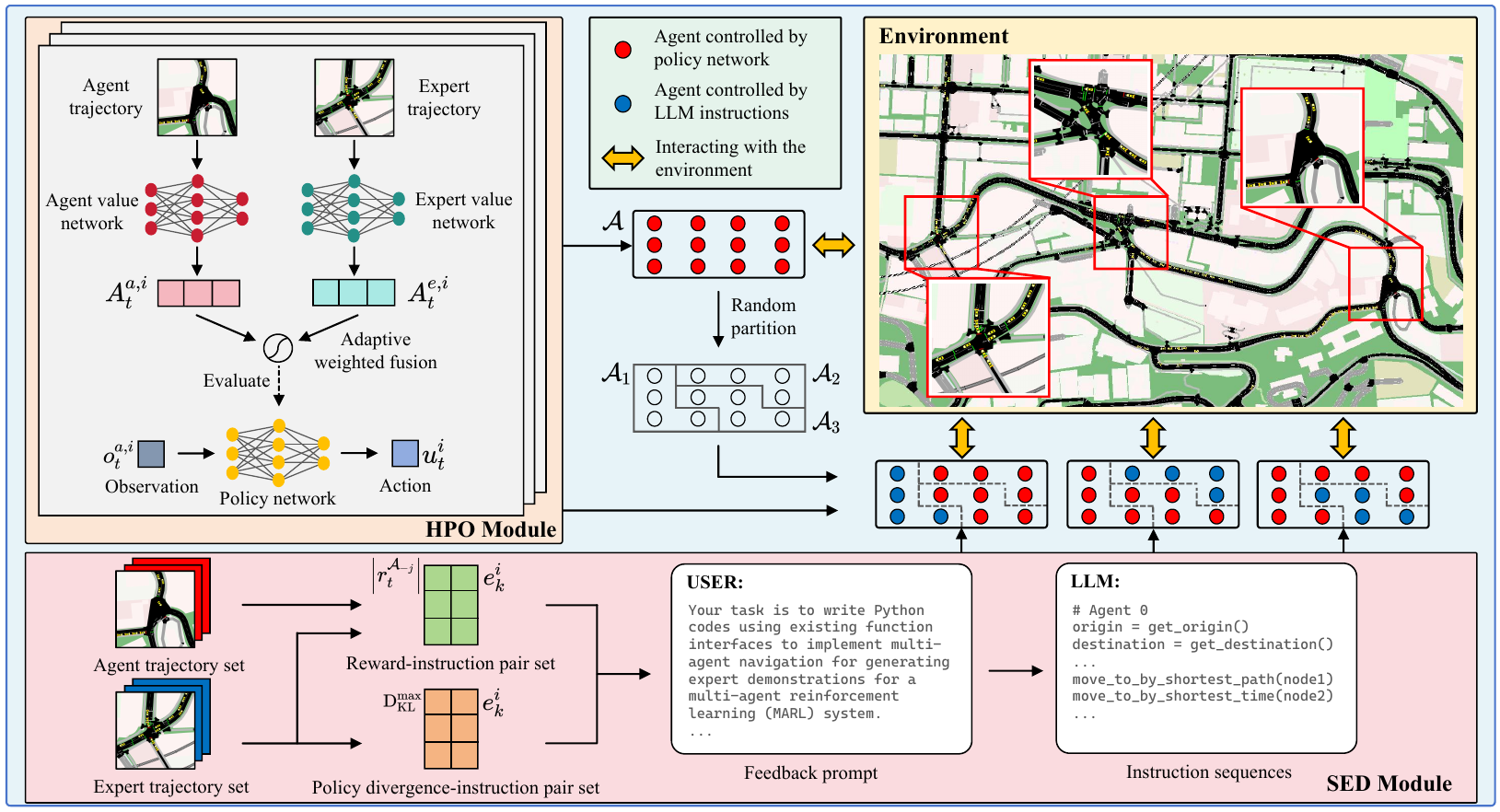}
    \caption{An overview of RELED framework.} 
    \label{fig:Overview.}
\end{figure*}

\section{RELED Framework}
\label{Sec: RELED Framework}
\subsection{System Model}

\textbf{Decentralized Partially Observable Markov Process.} 
In most MAS application scenarios, agents collaboratively accomplish tasks in a shared environment while only receiving partial local observations. As a result, MARL is formulated as a decentralized partially observable Markov process (Dec-POMDP)~\cite{oliehoek2016concise}, which is formally defined by the tuple $\left \langle \mathcal{A}, \mathcal{S}, \boldsymbol{\mathcal{U}}, P, \mathcal{R},\Omega ,\mathcal{O}, \rho_0, \gamma \right \rangle$. Specifically, $\mathcal{A} =\left \{ 1, \dots, n \right \} $ represents the set of $n$ agents, and $\mathcal{S}$ denotes the state space. The joint action space is given by $\boldsymbol{\mathcal{U}} = \prod_{i=1}^n \mathcal{U}^i$, where $\mathcal{U}^i$ denotes the action space of agent $i$. The state transition function $P: \mathcal{S} \times \boldsymbol{\mathcal{U}} \times \mathcal{S} \to [0, 1]$ defines the probability of transitioning to the next state given the current state and joint action. The reward function $\mathcal{R}=\left \{ R^i \right \}_{i\in \mathcal{A} }$, where $R^i: \mathcal{S} \times \boldsymbol{\mathcal{U}} \times \mathcal{S} \to \mathbb{R}$ assigns a reward to each agent. $\Omega$ denotes the observation space, and $\mathcal{O}=\left \{ O^i \right \}_{i\in \mathcal{A} }$, with $O^i: \mathcal{S} \times \mathcal{U}^i \to \Omega$ as the observation function for agent $i$. The initial state distribution $\rho_0: \mathcal{S} \to [0, 1]$ defines the probability of each initial state. The discount factor $\gamma \in [0, 1]$ determines the relative importance of future rewards.

\textbf{Interaction Process and Agent Trajectory.} 
In Dec-POMDP, the policy $\pi^i: \Omega \times \mathcal{U}^i \to [0, 1]$ denotes the probability distribution over actions for agent $i$. An episode begins with the environment sampling an initial state $s_0 \sim \rho_0$. At each time step $t$, agent $i$ receives a local observation $o_t^i \in \Omega$ according to its observation function $O^i(s_t)$. Each agent then selects an action $u^i_t \sim \pi^i \left ( \cdot \mid o^i_t \right )$, forming the joint action $\mathbf{u}_t = (u_t^1, \dots, u_t^n)$. The environment transitions to $s_{t+1} \sim P(\cdot \mid s_t, \mathbf{u}_t)$ and provides each agent with a reward $r^i_t=R^i(s_t, \mathbf{u}_t, s_{t+1})$. The process continues, with each agent receiving an observation-action-reward tuple at each time step, which is formalized as follows:
\begin{equation}
    \label{AgentPolicyRollout}
    \boldsymbol{\tau }^a \sim \mathrm{Env}(\boldsymbol{\pi  } \mid s_0,P),
\end{equation}
where $\boldsymbol{\tau}^a=\left \{ \tau^i\mid i\in \mathcal{A}  \right \} $ denotes the agent trajectory set, $\tau^i=\left(o^i_0, u^i_0, r^i_0, o^i_1, u^i_1, r^i_1, \dots\right)$ denotes the observation trajectory obtained by agent $i$, and $\mathrm{Env}(\cdot)$ denotes the environment. 

\textbf{Objective Function.} 
The objective for each agent is to maximize the expected return, defined as the expected cumulative discounted reward over its trajectory:
\begin{equation}
\label{Agent Objective Function}
    J^i \left (\boldsymbol{\pi}  \right ) = \mathbb{E}_{\rho_0,\boldsymbol{\pi},P} \left [ \sum_{t=0}^{\infty }\gamma ^t r^i_t \right ] ,
\end{equation}
where $\boldsymbol{\pi}\left ( \cdot \mid \mathbf{o} _t \right )= {\textstyle \prod_{i\in {\mathcal{A}}}} \pi^i\left ( \cdot \mid o^i_t \right )$ denotes the joint policy of all agents. The overall objective is to maximize the sum of individual agent objectives:
\begin{align}
J(\boldsymbol{\pi}  ) & =  \sum_{i \in \mathcal{A} } {J^i \left (\boldsymbol{\pi}  \right ) }  \notag \\
& = \mathbb{E}_{\rho_0,\boldsymbol{\pi},P} \left [ \sum_{t=0}^{\infty } \sum_{i \in \mathcal{A} }\gamma^t r^i_t \right ].
\end{align}

\subsection{System Overview}

In this section, we propose RELED, a scalable policy optimization framework that integrates LLM-driven expert demonstrations with agent-driven autonomous exploration. The framework consists of two key modules: an Stationarity-Aware Expert Demonstration (SED) module and a Hybrid Expert-Agent Policy Optimization (HPO) module, as outlined below:


\begin{itemize}
\item To mitigate the intrinsic non-stationarity arising from simultaneous policy updates in MARL systems, we propose the Stationarity-Aware Expert Demonstration (SED) module (see Section~\ref{Sec: Expert Demonstration Mechanism}). Based on a theoretically derived upper bound on environmental non-stationarity, the SED module quantitatively estimates the impact of policy changes and leverages LLMs to generate targeted expert demonstration trajectories for each agent. By incorporating both the reward volatility index and the policy divergence index, which are derived from our theoretical analysis, as feedback, the generated expert demonstrations can effectively stabilize the environment and accelerate policy convergence for all agents.

\item To address the limited coverage of the state-action space and the poor generalization associated with pure imitation learning from expert demonstrations, we propose the Hybrid Expert-Agent Policy Optimization (HPO) module (see Section~\ref{Sec: Hybrid policy optimization mechanism}). The HPO module separately evaluates policy losses on both expert-generated and agent-generated trajectories, and integrates them through a dynamically adjusted weighting scheme. This adaptive mechanism enables agents to gradually shift from expert-driven imitation to autonomous exploration as training progresses, accelerating policy convergence and mitigating suboptimal solutions.
\end{itemize}

As illustrated in Figure~\ref{fig:Overview.}, during each iteration of expert demonstration generation, the agent set $\mathcal{A}$ is randomly partitioned into $m$ disjoint subsets. Agents within each subset interact with the environment by following instruction sequences generated by the SED module, while the remaining agents execute their current policies. Within the SED module, reward-instruction and policy divergence-instruction pair sets are constructed by analyzing the resulting trajectory sets. These pair sets then provide feedback to the LLM, enabling iterative refinement of the instruction sequences. The HPO module adopts a fully decentralized training framework, where each agent independently optimizes its policy by leveraging both its own trajectories and those generated via instruction sequences. The agent and expert advantages, $A^{a,i}_t$ and $A^{e,i}_t$, are computed based on their respective value functions and are used to define separate policy losses. An adaptive weighted fusion mechanism dynamically adjusts the contributions of agent and expert losses according to the similarity between agent and expert trajectories, achieving hybrid policy optimization.

\subsection{Theoretical Analysis of Non-Stationarity in MARL System}
\label{Sec: Non-stationary boundary}

In this section, we present a theoretical analysis of environmental non-stationarity in MARL systems, serving as the foundation for the SED module.

A fundamental challenge in MARL arises from the inherent non-stationarity of the environment, induced by the simultaneous policy updates of multiple agents. As each agent updates its policy, the environment perceived by other agents changes dynamically, leading to biased estimates of its objective functions. This bias stems from the difficulty in distinguishing whether reward fluctuations originate from an agent’s own actions or from the evolving policies of other agents. Therefore, it is essential to quantify the impact of each agent's policy changes on the value estimations of other agents. Such quantification enables the targeted refinement of LLM-generated expert demonstrations, thereby enhancing the stability and performance of MARL systems.

However, directly assessing the impact of each agent’s policy changes on the objectives of other agents results in sample complexity and computational overhead growing multiplicatively with the number of agents. To address this problem, we partition the agent set $\mathcal{A}$ into $m$ disjoint subsets $\left \{ \mathcal{A}_1, \ldots, \mathcal{A}_m \right \} $, and focus on the joint objective function of the other agents when the policies of agents within a specific subset $\mathcal{A}_j$ are changed. Specifically, for a given subset of agents $\mathcal{A}_j$, we first define the joint objective of the external agents as follows:
\begin{align}
J^{{\mathcal{A}_{-j}}}(\boldsymbol{\pi}) & =  \sum_{i \notin \mathcal{A}_j } {J^i \left (\boldsymbol{\pi}  \right ) } \notag \\
& = \mathbb{E}_{\rho_0,\boldsymbol{\pi},P} \left[ \sum_{t=0}^{\infty }  \gamma^t r^{\mathcal{A}_{-j}}_t \right],
\end{align}
where $\mathcal{A}_{-j}= \mathcal{A}\setminus \mathcal{A}_j$ denotes the external agents of subset $\mathcal{A}_j$, i.e., all agents not in $\mathcal{A}_j$, and $r^{\mathcal{A}_{-j}}_t =\sum_{i \notin \mathcal{A}_j } r^i_t $ denotes the sum of rewards obtained by external agents of subset $\mathcal{A}_j$ at time step $t$. We then derive the variation bounds of the external agents’ objective function under different policy configurations of an agent subset, thereby quantifying the resulting environmental non-stationarity, as formalized in the following theorem:
\begin{theorem} {\bf{(Performance Bound on Group-Level Non-Stationarity)}} \label{the:1}
Consider two arbitrary sets of joint policies for an agent subset $\mathcal{A}_j$:
\begin{align*}
\boldsymbol{\widehat{\pi}}^{\mathcal{A}_j}\left ( \cdot \mid \mathbf{o}^j_t  \right ) &= {\textstyle \prod_{i\in {\mathcal{A}_j}}} \boldsymbol{\widehat{\pi}}^i\left ( \cdot \mid o^i_t \right ), \\
\boldsymbol{\widetilde{\pi}}^{\mathcal{A}_j}\left ( \cdot \mid \mathbf{o}^j_t  \right ) &= {\textstyle \prod_{i\in {\mathcal{A}_j}}} \boldsymbol{\widetilde{\pi}}^i\left ( \cdot \mid o^i_t \right ),
\end{align*}
where $\mathbf{o}^j_t=\left ( o^k_t \right )_{k\in \mathcal{A}_{j}}$ denotes the joint observation of the agent subset $\mathcal{A}_j$. Let $\boldsymbol{\pi}^{{\mathcal{A}_{-j}}} = \left\{\, \pi^i \mid i \in \mathcal{A} \setminus {\mathcal{A}_j} \, \right\}$ denote the set of policies for agents external to the subset $\mathcal{A}_j$, and define the joint policies $\widehat{\boldsymbol{\pi}}$ and $\widetilde{\boldsymbol{\pi}}$ as:

\begin{align*}
\widehat{\boldsymbol{\pi}} \left ( \cdot \mid \mathbf{o}_t \right ) &=\widehat{\boldsymbol{\pi}}^{{\mathcal{A}_j}} \left ( \cdot \mid \mathbf{o}^j_t \right )  {\textstyle \prod_{k \notin  {\mathcal{A}_j}}\pi ^k\left ( \cdot \mid o^k_t \right )  }, \\ 
\widetilde{\boldsymbol{\pi}} \left ( \cdot \mid \mathbf{o}_t \right ) &=\widetilde{\boldsymbol{\pi}}^{{\mathcal{A}_j}} \left ( \cdot \mid \mathbf{o}^j_t \right )  {\textstyle \prod_{k \notin  {\mathcal{A}_j}}\pi^k\left ( \cdot \mid o^k_t \right )  }.
\end{align*}
Then, the following bound holds:
\small
\begin{align}
& J^{\mathcal{A}_{-j}} \left (\widehat{\boldsymbol{\pi}} \right ) 
- J^{\mathcal{A}_{-j}}  \left (\widetilde{\boldsymbol{\pi}}\right ) \notag \\
& \le  \frac{\sqrt{2}}{\left ( 1-\gamma \right )^2 } 
\max_{\boldsymbol{\widehat{\tau}},\boldsymbol{\widetilde{\tau}}} 
\left | r^{\mathcal{A}_{-j}}_t  \right | \sum_{k \in  {\mathcal{A}_j}}  {
\mathrm{D}^\mathrm{max}_\mathrm{KL}\left ( \widehat{\pi}^k \parallel \widetilde{\pi}^k \right )}, 
\end{align}
where $\boldsymbol{\widehat{\tau}}$ and $\boldsymbol{\widetilde{\tau}}$ denote trajectory rollouts of the joint policies $\boldsymbol{\widehat{\pi}}$ and $\boldsymbol{\widetilde{\pi}}$, respectively, and $\mathrm{D}^\mathrm{max}_\mathrm{KL}\left ( \widehat{\pi}^k \parallel \widetilde{\pi}^k \right )=\max_{o^k_t} {\mathrm{D}_\mathrm{KL}\left ( \widehat{\pi}^k \parallel \widetilde{\pi}^k \right )} \left [ o^k_t  \right ]$ denotes the maximal Kullback-Leibler (KL) divergence between policies $\widehat{\pi}^k$ and $\widetilde{\pi}^k$.
\begin{proof}
We build upon intermediate results from prior work on performance-difference bounds, extend them to partially observable multi-agent settings, and further derive a broader bound to facilitate subsequent estimation in practice. Specifically, we first assume $\gamma f\left ( s' \right ) \equiv f\left ( s \right )$ in Theorem 1 of Achiam et al. \cite{achiam2017constrained}, and extend it to joint policies $\widehat{\boldsymbol{\pi}}$ and $\widetilde{\boldsymbol{\pi}}$, which yields:
\begin{align}
\label{eq:Performance bound}
& J^{\mathcal{A}_j} \left (\widehat{\boldsymbol{\pi}} \right ) 
- J^{\mathcal{A}_j}  \left (\widetilde{\boldsymbol{\pi}}\right ) \notag \\
& \le \frac{1}{1-\gamma } \mathbb{E}_{\substack{s_t\sim d^{\widetilde{\boldsymbol{\pi}}} \\ \mathbf{u}_t \sim \widetilde{\boldsymbol{\pi}} \\ s_{t+1}\sim P}}\left [\left ( \frac{\widehat{\boldsymbol{\pi}} \left ( \mathbf{u}_t \mid \mathbf{o}_t  \right ) }{\widetilde{\boldsymbol{\pi}}  \left ( \mathbf{u}_t \mid \mathbf{o}_t \right )} -1 \right )r^{\mathcal{A}_{-j}}_t  \right ] \notag \\
&+ \frac{2\gamma }{\left ( 1-\gamma  \right )^2 } \max_{s_t} \left | \mathbb{E}_{\substack{\mathbf{u}_t\sim \boldsymbol{\widehat{\pi}} \\ s_{t+1}\sim P}} \left [ r^{\mathcal{A}_{-j}}_t \right ]  \right |   \mathbb{E} _{\substack{s_t\sim d^{\widetilde{\boldsymbol{\pi}}} }}\left [ \mathrm{D}_\mathrm{TV}\left ( \widehat{\boldsymbol{\pi}} \parallel   \widetilde{\boldsymbol{\pi}}\right )\left [ \mathbf{o}_t  \right ]   \right ], 
\end{align}
where $d^{\widetilde{\boldsymbol{\pi}}}$ denotes the discounted future state distribution under the joint policy $\widetilde{\boldsymbol{\pi}}$:
\begin{equation}
    d^{\widetilde{\boldsymbol{\pi}}}\left ( s \right ) =\left ( 1-\gamma  \right ) \sum_{t=0}^{\infty }\mathrm{Pr}\left ( s_t=s \mid \rho_0,P,\widetilde{\boldsymbol{\pi}}\right ),
\end{equation}
and $\mathrm{D}_\mathrm{TV}\left ( \widehat{\boldsymbol{\pi}} \parallel   \widetilde{\boldsymbol{\pi}}\right )\left [ \mathbf{o}_t  \right ]$ denotes the total variation distance between $ \widehat{\boldsymbol{\pi}}$ and $\widetilde{\boldsymbol{\pi}}$ at $\mathbf{o}_t$:
\begin{equation}
    \mathrm{D}_\mathrm{TV}\left ( \widehat{\boldsymbol{\pi}} \parallel   \widetilde{\boldsymbol{\pi}}\right )\left [ \mathbf{o}_t  \right ]=\frac{1}{2}\sum_{\mathbf{u}_t}\left | \widehat{\boldsymbol{\pi}}\left (  \mathbf{u}_t\mid \mathbf{o}_t  \right ) -\widetilde{\boldsymbol{\pi}}\left (  \mathbf{u}_t\mid \mathbf{o}_t  \right ) \right |.
\end{equation}
We bound the first term on the right side of Eqn.~\ref{eq:Performance bound}:
\begin{align}
\label{eq:First term of Performance bound}
& \frac{1}{1-\gamma } \mathbb{E}_{\substack{s_t\sim d^{\widetilde{\boldsymbol{\pi}}} \\  \mathbf{u}_t \sim \widetilde{\boldsymbol{\pi}} \\ s_{t+1}\sim P}}\left [\left ( \frac{\widehat{\boldsymbol{\pi}} \left ( \mathbf{u}_t \mid \mathbf{o}_t  \right ) }{\widetilde{\boldsymbol{\pi}}  \left ( \mathbf{u}_t \mid \mathbf{o}_t \right )} -1 \right )r^{\mathcal{A}_{-j}}_t  \right ] \notag \\
& = \frac{1}{1-\gamma } \mathbb{E}_{\substack{s_t\sim d^{\widetilde{\boldsymbol{\pi}}}\\s_{t+1}\sim P}} \left [ \sum_{\mathbf{u}_t} \left [ \widehat{\boldsymbol{\pi}} \left ( \mathbf{u}_t \mid \mathbf{o}_t  \right ) -\widetilde{\boldsymbol{\pi}}  \left ( \mathbf{u}_t \mid \mathbf{o}_t \right ) \right ]  r^{\mathcal{A}_{-j}}_t  \right ] \notag \\
& \le \frac{1}{1-\gamma } \max_{\mathbf{o}_t} \sum_{\mathbf{u}_t} \left | \widehat{\boldsymbol{\pi}} \left ( \mathbf{u}_t \mid \mathbf{o}_t  \right ) -\widetilde{\boldsymbol{\pi}}  \left ( \mathbf{u}_t \mid \mathbf{o}_t \right ) \right | \max_{\widehat{\boldsymbol{\tau}},\widetilde{\boldsymbol{\tau }}}\left | r^{\mathcal{A}_{-j}}_t \right | \notag \\
& = \frac{2}{1-\gamma } \max_{\mathbf{o}_t}  \mathrm{D}_\mathrm{TV}\left ( \widehat{\boldsymbol{\pi}} \parallel   \widetilde{\boldsymbol{\pi}}    \right ) \left [ \mathbf{o}_t  \right ]\max_{\widehat{\boldsymbol{\tau}},\widetilde{\boldsymbol{\tau }}}\left | r^{\mathcal{A}_{-j}}_t \right | \notag \\
& \le \frac{\sqrt{2} }{1-\gamma } \max_{\mathbf{o}_t}  \mathrm{D}_\mathrm{KL}\left ( \widehat{\boldsymbol{\pi}} \parallel   \widetilde{\boldsymbol{\pi}}    \right ) \left [ \mathbf{o}_t  \right ]\max_{\widehat{\boldsymbol{\tau}},\widetilde{\boldsymbol{\tau }}}\left | r^{\mathcal{A}_{-j}}_t \right | ,
\end{align}
where $\mathrm{D}_\mathrm{KL}\left ( \widehat{\boldsymbol{\pi}} \parallel   \widetilde{\boldsymbol{\pi}}    \right ) \left [ \mathbf{o}_t  \right ] $ denotes the KL divergence between $\widehat{\boldsymbol{\pi}}$ and $\widetilde{\boldsymbol{\pi}}$ at $o_t$: 
\begin{equation}
    \mathrm{D}_\mathrm{KL}\left ( \widehat{\boldsymbol{\pi}} \parallel   \widetilde{\boldsymbol{\pi}}\right )\left [ \mathbf{o}_t  \right ]=\mathbb{E} _{\mathbf{u}_t \sim\widehat{\boldsymbol{\pi}}}\left [ \log{\frac{ \widehat{\boldsymbol{\pi}} \left ( \mathbf{u}_t \mid \mathbf{o}_t  \right )}{ \widetilde{\boldsymbol{\pi}} \left ( \mathbf{u}_t \mid \mathbf{o}_t  \right )} } \right ] .
\end{equation}
Then, we bound the second term in Eqn.~\ref{eq:Performance bound}:
\begin{align}
\label{eq:Second term of Performance bound}
& \frac{2\gamma }{\left ( 1-\gamma  \right )^2 } \max_{s_t} \left | \mathbb{E}_{\substack{\mathbf{u}_t\sim \boldsymbol{\widehat{\pi}} \\ s_{t+1}\sim P}} \left [r^{\mathcal{A}_{-j}}_t \right ]  \right |   \mathbb{E} _{\substack{s_t\sim d^{\widetilde{\boldsymbol{\pi}}} }}\left [ \mathrm{D}_\mathrm{TV}\left ( \widehat{\boldsymbol{\pi}} \parallel   \widetilde{\boldsymbol{\pi}}\right )\left [ \mathbf{o}_t  \right ]  \right ] \notag \\
& \le \frac{2\gamma }{ \left ( 1-\gamma \right )^2 } \max_{\boldsymbol{\widehat{\tau}}} \left | r^{\mathcal{A}_{-j}}_t  \right | \max_{\mathbf{o}_t }\mathrm{D}_\mathrm{TV}\left ( \widehat{\boldsymbol{\pi}} \parallel   \widetilde{\boldsymbol{\pi}}    \right ) \left [ \mathbf{o}_t  \right ]\notag \\
& \le \frac{\sqrt{2} \gamma }{ \left ( 1-\gamma \right )^2 } \max_{\boldsymbol{\widehat{\tau}}} \left | r^{\mathcal{A}_{-j}}_t \right | \max_{\mathbf{o}_t }\mathrm{D}_\mathrm{KL}\left ( \widehat{\boldsymbol{\pi}} \parallel   \widetilde{\boldsymbol{\pi}}    \right )\left [ \mathbf{o}_t  \right ] .
\end{align}
By combining Eqn.~\ref{eq:First term of Performance bound} and Eqn.~\ref{eq:Second term of Performance bound} and comparing with Eqn.~\ref{eq:Performance bound}, we obtain:
\begin{align}
& J^{\mathcal{A}_j} \left (\widehat{\boldsymbol{\pi}} \right ) 
- J^{\mathcal{A}_j}  \left (\widetilde{\boldsymbol{\pi}}\right ) \notag \\
& \le \frac{\sqrt{2}}{ \left ( 1-\gamma \right )^2 } \max_{\boldsymbol{\widehat{\tau}},\boldsymbol{\widetilde{\tau}}} \left | r^{\mathcal{A}_{-j}}_t  \right | \max_{\mathbf{o}_t }\mathrm{D}_\mathrm{KL}\left ( \widehat{\boldsymbol{\pi}} \parallel   \widetilde{\boldsymbol{\pi}}    \right )\left [ \mathbf{o}_t  \right ] \notag \\
& \le \frac{\sqrt{2}}{ \left ( 1-\gamma \right )^2 } \max_{\boldsymbol{\widehat{\tau}},\boldsymbol{\widetilde{\tau}}} \left | r^{\mathcal{A}_{-j}}_t  \right | \sum_{k \in  {\mathcal{A}_j}}{\mathrm{D}^\mathrm{max}_\mathrm{KL}\left ( \widehat{\pi}^k \parallel \widetilde{\pi}^k \right )} 
\end{align}
This concludes the proof.
\end{proof}
\end{theorem}

\textbf{Remark.}
Theorem~\ref{the:1} decomposes the impact of non-stationarity induced by policy changes within an agent subset $\mathcal{A}_j$ into two components: (i) the maximal fluctuation in the cumulative rewards of external agents, quantified by the \emph{reward volatility index} $\max_{\boldsymbol{\widehat{\tau}},\boldsymbol{\widetilde{\tau}}} \left | r^{\mathcal{A}_{-j}}_t \right |$, and (ii) the aggregate maximal shift in the policy distributions within the subset, quantified by the \emph{policy divergence index} $\sum_{k \in {\mathcal{A}j}} {\mathrm{D}^\mathrm{max}_\mathrm{KL}\left ( \widehat{\pi}^k \parallel \widetilde{\pi}^k \right )}$. Theorem~\ref{the:1} demonstrates that even minor deviations in individual agent policies can collectively exacerbate instability in the objective functions of other agents. This highlights the importance of controlling individual policy changes to ensure stability in MARL systems.

Building on Theorem~\ref{the:1}, we derive an immediate corollary that bounds the agent-level non-stationarity.

\begin{corollary} {\bf{(Performance Bound on Agent-Level Non-Stationarity)}} \label{cor:1}
Consider two arbitrary policies $\widehat{\pi}^i$ and $\widetilde{\pi}^i$ for agent $i$, and a policy sets for $\mathcal{A}\setminus{\left\{\,i \right\}}$:
\begin{equation}
    \boldsymbol{\pi}^{-i} = \left \{ \pi^1,\dots ,\pi^{i-1}, \pi^{i+1},\dots ,\pi^{n} \right \}.
\end{equation}
Define the joint policies $\widehat{\boldsymbol{\pi}}$ and $\widetilde{\boldsymbol{\pi}}$ as:
\begin{align*}
\widehat{\boldsymbol{\pi}} \left ( \cdot \mid \mathbf{o}_t \right ) &=\widehat{\pi} ^i \left ( \cdot \mid o^i_t \right )  {\textstyle \prod_{j\ne i}\pi ^j\left ( \cdot \mid o^j_t \right )  }, \\ 
\widetilde{\boldsymbol{\pi}} \left ( \cdot \mid \mathbf{o}_t \right ) &=\widetilde{\pi} ^i \left ( \cdot \mid o^i_t \right )  {\textstyle \prod_{j\ne i}\pi ^j\left ( \cdot \mid o^j_t \right )  }.
\end{align*}
Then, the following inequality holds:
\begin{align}
& J^i \left (\widehat{\boldsymbol{\pi}} \right ) - J^i \left ( \widetilde{\boldsymbol{\pi}}\right ) \notag \\
& \le  \frac{\sqrt{2}}{\left ( 1-\gamma \right )^2 } \max_{\boldsymbol{\widehat{\tau}},\boldsymbol{\widetilde{\tau}}} \left |\sum_{j \ne i}{r^j_t}\right| \max_{o^i_t} {\mathrm{D}_\mathrm{KL}\left ( \widehat{\pi}^i \parallel \widetilde{\pi}^i \right )} \left [ o^i_t  \right ].
\end{align}
\begin{proof}
The result follows from Theorem~\ref{the:1} by setting $\mathcal{A}_j=\left \{ i \right \} $.
\end{proof}
\end{corollary}

\textbf{Remark.}
Corollary~\ref{cor:1} establishes an upper bound on the impact of an agent's policy change on the objective functions of other agents. However, applying this upper bound necessitates individual evaluation for each agent, resulting in a total of $n$ estimations and a computational complexity that scales linearly with the number of agents. In contrast, Theorem~\ref{the:1} introduces a subset-based bound, reducing the required evaluations to $m$. Therefore, the bound estimation based on Theorem~\ref{the:1} reduces computational cost and thus is more scalable and practical for MARL systems.

\subsection{Stationarity-Aware Expert Demonstration Module}
\label{Sec: Expert Demonstration Mechanism}

In this section, we propose the SED module, which leverages LLMs to generate expert demonstrations for each agent. The SED module incorporates both the reward volatility index and the policy divergence index, derived from the theoretical results in Section~\ref{Sec: Non-stationary boundary}, as feedback to iteratively query LLMs for demonstrations with high expected returns and enhanced environmental stability.

Conventional expert demonstrations for MARL typically relies on rule-crafted or human-curated demonstrations~\cite{yu2024beyond,weiwei2023expert}, which are costly to obtain and tailored to specific task scenarios. In contrast, LLMs, with their strong contextual reasoning capabilities and extensive prior knowledge, have shown great potential in generating customized expert demonstrations for MARL agents~\cite{wei2025plangenllms,zhou2024isr}. Furthermore, by continuously providing environmental feedback to the LLM, the generated demonstrations can be dynamically adapted for each agent, thereby producing high-quality demonstration samples and further accelerating the convergence of agent policies.

Therefore, we employ an LLM-driven expert demonstration generation mechanism in the SED module. For clarity, we denote the LLM as $\Pi \left ( d \right ) $, where $d$ is the prompt input. The objective of the LLM is to generate an instruction sequence for each agent, formalized as $\mathcal{E} =\left \{ e^i_k\mid i\in \mathcal{A},k=0,1,\dots \right \} $. Each instruction sequence $\mathbf{e}^i$ comprises ordered task steps that guide agent $i$ to interact with the environment. To ensure executability, each instruction $e^i_j$ is selected from a predefined set $\mathcal{C}$. The environment then maps each instruction to an executable action sequence via an execution function $f: \mathcal{C} \to \boldsymbol{\mathcal{U}}^l$, defined as:
\begin{equation}
    f\left ( e^i_k \right ) =\left ( u^i_{t_k},u^i_{t_k+1},\dots ,u^i_{t_k+l^i_k-1} \right ) 
\end{equation}
where $0< l^i_k\le l$ denotes the number of timesteps required for agent $i$ to complete instruction $e^i_k$. In implementation, $f$ corresponds to a set of programmatic functions that realize the semantics of instructions in $\mathcal{C}$. For example, in vehicle routing, an instruction such as \texttt{move\_to\_by\_shortest\_path(node\_id)} is operationalized by invoking a function that directs the agent to the specified node via the shortest path. Further implementation details are provided in Section~\ref{Sec: Implementation}. 

In Theorem~\ref{the:1}, we quantitatively characterize the impact of policy changes within a subset of agents on the objective functions of other agents. This provides a theoretical bound for analyzing the non-stationarity induced in the environment when that subset updates its policy by imitating expert demonstrations. Accordingly, the SED module employs the reward volatility index and policy divergence index, both derived from this bound, as two feedback metrics for the LLM. Specifically, let $\widetilde{\boldsymbol{\pi}}$ and $\widehat{\boldsymbol{\pi}}$ denote the joint policies of the agent subset $\mathcal{A}_j$ before and after the update, respectively. By minimizing the upper bound established in Theorem~\ref{the:1}, the SED module constrains the fluctuations in the objective functions of external agents caused by policy shifts within $\mathcal{A}_j$. To this end,  during each iteration, the SED module randomly partitions the agent set $\mathcal{A}$ into $m$ disjoint subsets $\{\mathcal{A}_1, \ldots, \mathcal{A}_m\}$ to comprehensively evaluate the non-stationarity induced by different agent combinations. For each subset $\mathcal{A}_j$, agents in the subset selects actions according to the instruction sequence $\mathcal{E}^{{\mathcal{A}_j}} = \left \{ \boldsymbol{e}^i \mid  i \in \mathcal{A}_j,k=0,1,\dots  \right \}$ generated by the LLM, executed via function $f$, while the remaining agents follow their respective policies. All agents interact with the environment to generate the trajectory set $\boldsymbol{\widehat{\tau}}$, formalized as:
\begin{equation}
    \label{AgentExpertPolicyRollout}
    \boldsymbol{\widehat{\tau}}=\boldsymbol{\tau}^{e,j} \sim \mathrm{Env}\left( \mathcal{E}^{{\mathcal{A}_j}}, \boldsymbol{\pi}^{\mathcal{A}_{-j}} \mid s_0,P,f\right),
\end{equation}
where $\boldsymbol{\pi}^{\mathcal{A}{-j}} = {\textstyle \prod_{i \notin {\mathcal{A}_j}}} \pi^i$ denotes the joint policy of agents outside $\mathcal{A}_j$. For comparison, the agent trajectory set $\boldsymbol{\widetilde{\tau}}$ is obtained via rollout under the current joint policy (i.e., $\boldsymbol{\widetilde{\tau}}=\boldsymbol{\tau}^a$).

To minimize the reward volatility index, expert demonstrations should ensure that updating the policy of the agent subset $\mathcal{A}_j$ results in a decrease in the reward fluctuations of external agents, which can be formalized as the following constraint:
\begin{equation}
   \max \left \{ \left | \min_{\boldsymbol{\widehat{\tau}}}r^{\mathcal{A}_{-j}}_t \right | ,\max_{\boldsymbol{\widehat{\tau}}}  r^{\mathcal{A}_{-j}}_t  \right \} \le  \max_{\boldsymbol{\widetilde{\tau}}} \left | r^{\mathcal{A}_{-j}}_t  \right |.
\end{equation}
However, in cooperative settings with coupled or global rewards, multiple agents receive identical rewards. In these scenarios, policy improvement within the agent subset $\mathcal{A}_j$ can inevitably cause an increase in $\max_{\boldsymbol{\widehat{\tau}}} r^{\mathcal{A}_{-j}}_t$. Therefore, the SED module focuses on constraining the term $\left| \min_{\boldsymbol{\widehat{\tau}}} r^{\mathcal{A}_{-j}}_t \right|$ within the range $\max_{\boldsymbol{\widetilde{\tau}}} \left | r^{\mathcal{A}_{-j}}_t \right |$, thereby optimizing worst-case reward fluctuations. Specifically, the SED module constructs a timestep set $\mathcal{T}^j$, consisting of all timesteps in the expert demonstration trajectory set $\boldsymbol{\widehat{\tau}}$ where the external agent’s reward is lower than the maximum absolute reward in the agent trajectory set $\boldsymbol{\widetilde{\tau}}$:
\begin{equation}
    \mathcal{T}^j=\left \{ t\mid  r^{\mathcal{A}_{-j}}_t  < -\max_{\boldsymbol{\widetilde{\tau}}} \left | r^{\mathcal{A}_{-j}}_t  \right | ,r^{\mathcal{A}_{-j}}_t\in \boldsymbol{\widehat{\tau}}\right \} ,
\end{equation}
The SED module further establishes the correspondence between the reward volatility index and LLM-generated instructions by constructing a reward-instruction pair set $\mathcal{R}^j$, which enables iterative feedback on the instruction sequence, as follows:
\begin{equation}
\label{Eq:reward volatility index}
    \mathcal{R}^j =\left \{ \left | r^{\mathcal{A}_{-j}}_t \right |,e^i_k\mid t\in \mathcal{T}^j,i\in {\mathcal{A}_j},l^i_{k-1}\le t< l^i_k\right \} .
\end{equation}

To minimize the policy divergence index, we estimate the KL divergence between the pre-update policy $\widetilde{\pi}^i$ and the post-update policy $\widehat{\pi}^i$ for each agent $i \in \mathcal{A}_j$. Given the trajectory set $\boldsymbol{\widehat{\tau}}$ and the current policy $\pi^i$, we approximate the updated policy $\widehat{\pi}^i$ as follows:
\begin{equation}
\label{ExpertPolicyApproximation}
   \widehat{\pi} ^i\left ( u^i\mid o^i_t \right )  =
\begin{cases}
    \delta \left ( u^i- u^i_t \right ) , & \exists  o^i_t,u^i_t \in \boldsymbol{\widehat{\tau}} \\
    \pi^i\left( u^i \mid o^i_t \right), & \text{otherwise}
\end{cases},
\end{equation}
where $\delta$ denotes the Dirac delta function. This approximation treats the observation-action pairs $ \left ( o^i_t,u^i_t \right )$ in the trajectory $\boldsymbol{\widehat{\tau}}$ as deterministic action outputs of the policy, while retaining the original policy distribution for all other observation inputs. In this way, the influence of expert demonstrations on the updated policy of agent $i$ is explicitly captured. Consequently, the KL divergence between the pre- and post-update policies for agent $i$ at an observation $o^i_t$ is given by:
\begin{align}
&\mathrm{D}_\mathrm{KL}\left ( \widehat{\pi}^i \parallel \widetilde{\pi}^i \right )[o^i_t] \notag \\
&=\mathbb{E} _{u^i_t\sim\widetilde{\pi}^i }\left [ \ln{\frac{\widehat{\pi}^i(u^i_t  \mid o^i_t)}{\pi^i(u^i_t \mid o^i_t)} }  \right ] \notag \\
&= \mathbb{E} _{ o^i_t,u^i_t  \in \boldsymbol{\widehat{\tau}}}\left [ \ln{\frac{1}{\pi^i(u^i_t \mid o^i_t)} }  \right ] + \mathbb{E} _{ o^i_t,u^i_t \notin \boldsymbol{\widehat{\tau}}}\left [ \ln{\frac{\pi^i(u^i_t \mid o^i_t)}{\pi^i(u^i_t \mid  o^i_t)} }  \right ] \notag \\
&=\mathbb{E} _{o^i_t,u^i_t \in \boldsymbol{\widehat{\tau}}}\left [ \ln{\frac{1}{\pi^i(u^i_t   \mid o^i_t)} }  \right ].
\end{align}
Therefore, the maximum policy divergence for agent $i$ can be expressed as:
\begin{equation}
    \mathrm{D}^\mathrm{max}_\mathrm{KL}\left ( \widehat{\pi}^i \parallel \widetilde{\pi}^i \right )=\max_{o^i_t,u^i_t  \in \boldsymbol{\widehat{\tau}}} {\ln{\frac{1}{\pi^i(u^i_t  \mid o^i_t)}}},
\end{equation}
and the timestep $t^*_i$ corresponding to the maximum KL divergence in the trajectory set $\boldsymbol{\widehat{\tau}}$ is given by:
\begin{equation}
    t^*_i= \operatorname*{argmax}_{t} {\ \ln{\frac{1}{\pi^i(u^i_t  \mid o^i_t)}}},
\end{equation}
where $(o^i_t,u^i_t) \in \boldsymbol{\widehat{\tau}}$. Based on the above, the SED module constructs the policy divergence-instruction pair set $\mathcal{K}^j$ for each agent subset $\mathcal{A}_j$ as:
\begin{equation}
    \label{Eq:policy divergence index}
    \mathcal{K}^j =\left \{ \mathrm{D}^\mathrm{max}_\mathrm{KL}\left ( \widehat{\pi}^i \parallel \widetilde{\pi}^i \right ),e^i_k\mid i\in {\mathcal{A}_j},l_{k-1}\le t^*_i< l_k\right \} 
\end{equation}

Through the above mechanism, the SED module constructs a reward-instruction pair set $\left \{ \mathcal{R}^j \right \} ^m_{j=1}$ and a policy divergence-instruction pair set $\left \{ \mathcal{K}^j \right \} ^m_{j=1}$ for each agent subset. The expert trajectory set are aggregated as $\boldsymbol{\tau}^e=\left \{ \tau^i\mid \tau^i\in \boldsymbol{\tau}^{e,j},i\in \mathcal{A}_j,j=1,\dots ,m  \right \}$. Accordingly, we design two LLM inputs $d$: (i) initial prompt $d_{\mathrm{init}}$, which provides a detailed description of the environment and reward structure to guide the LLM in generating task-targeted instruction sequences for each agent; and (ii) feedback prompt $d_{\mathrm{fb}}$, which incorporates observation information of the expert trajectory set $\boldsymbol{\tau}^e$  and instructs the LLM to revise instructions associated with significant reward fluctuations or excessive policy divergence, as identified via $\left \{ \mathcal{R}^j \right \} ^m_{j=1}$ and $\left \{ \mathcal{K}^j \right \} ^m_{j=1}$, respectively. A complete description of the prompts is provided in the Appendix A.

\begin{figure}[t]
  \centering
  \includegraphics[width=0.48\textwidth]{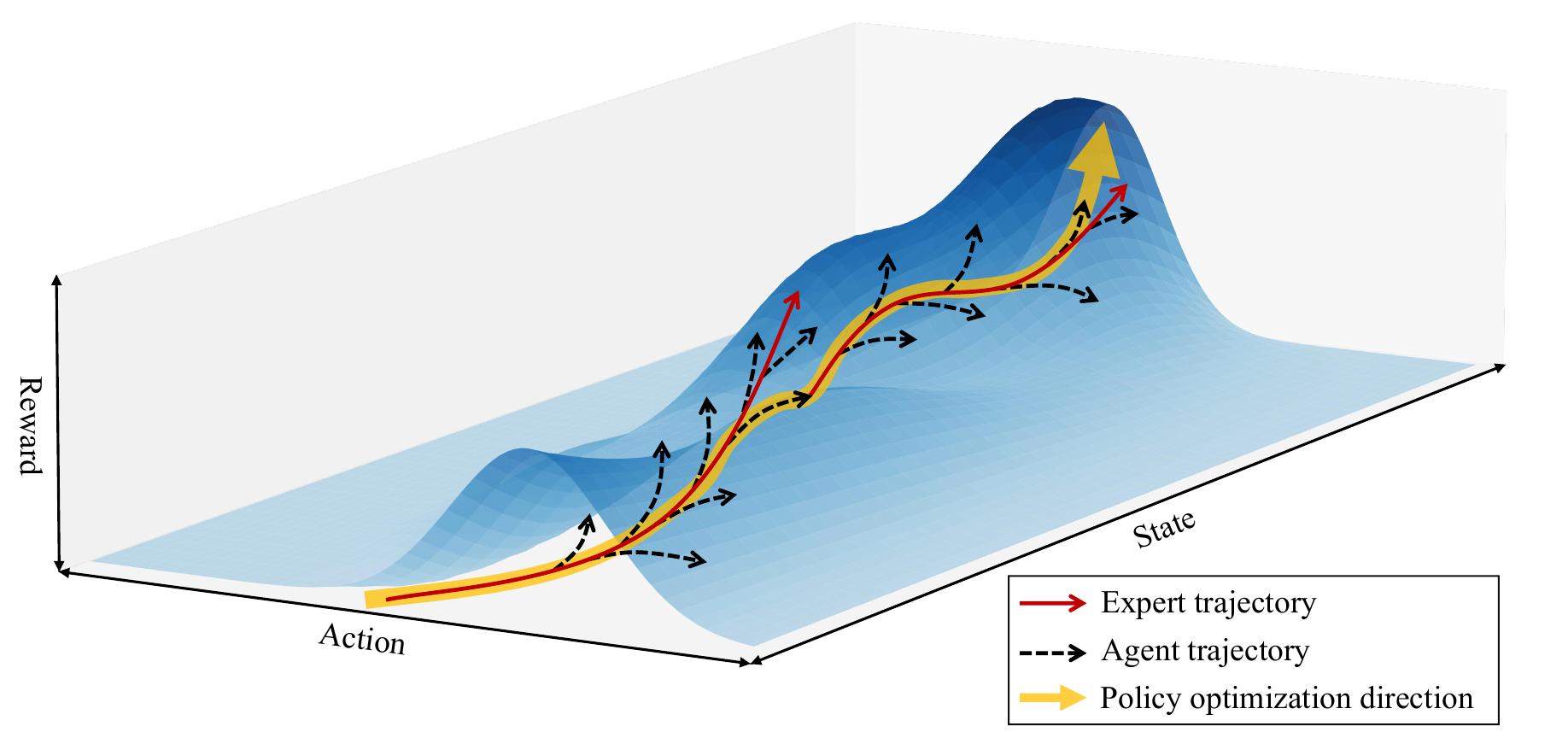}
  \caption{The hybrid policy optimization mechanism adaptively combines expert demonstration guidance (red solid line) and autonomous exploration (black dashed line), facilitating the gradual convergence of agent's policy towards the optimal solution.}
  \label{fig:2}
\end{figure}

\begin{algorithm}[t]
\caption{RELED Training Procedure }
\label{Alg:Training Procedure}
\begin{flushleft}
\textbf{Input}: $K$, $\Pi$, $d_{\mathrm{init}}$, $d_{\mathrm{fb}}$, $q$, $n$, $m$.

\textbf{Output}: $\boldsymbol{\pi}$.
\end{flushleft}
\begin{algorithmic}[1] 
\STATE {\bf{Random initialization:}} $\pi ^i$, $V^{a,i}$, $V^{e,i}$, $i=1,\dots,n$.
\STATE {\bf{Initialization:}} $\Pi$, $d=d_{\mathrm{init}}$.
\FOR{$k=1$ to $K$}
\STATE $\boldsymbol{\tau }^a \sim \text{Env}(\boldsymbol{\pi  } \mid s_0,P)$ \emph{// Sample agent trajectories}
\STATE \textit{/* Stationarity-Aware Expert Demonstration Module: */}
\IF{$k \bmod q = 0$}
\STATE Randomly partition $\mathcal{A}$ into $\{\mathcal{A}_1, \ldots, \mathcal{A}_m\}$.
\STATE $\mathcal{E}\sim \Pi \left ( d \right ) $ \emph{// Generate instruction sequences}
\STATE $d\gets d_{\mathrm{fb}}$, $\boldsymbol{\tau}^e \gets\emptyset $
\FOR{$j=1$ to $m$}
\STATE $\boldsymbol{\tau}^{e,j} \sim \mathrm{Env}\left( \mathcal{E}^{{\mathcal{A}_j}}, \boldsymbol{\pi}^{\mathcal{A}_{-j}} \mid s_0,P,f\right)$ \emph{// Sample expert trajectories}
\STATE Calculate $\mathcal{R}^j$ and $\mathcal{K}^j$ via Eqn.~\ref{Eq:reward volatility index} and~\ref{Eq:policy divergence index}.
\STATE $d\gets d \cup \left \{ \mathcal{R}^j,\mathcal{K}^j \right \}$
\STATE $\boldsymbol{\tau}^e \gets \boldsymbol{\tau}^e \cup \left \{ \tau^i\mid  \tau^i\in \boldsymbol{\tau}^{e,j},i\in \mathcal{A}_j  \right \} $
\ENDFOR
\STATE $d\gets d \cup \boldsymbol{\tau}^e$
\ENDIF
\STATE \textit{/* Hybrid Expert-Agent Policy Optimization Module: */}
\FOR{$i=1$ to $n$}
\STATE $\tau^{a,i} \gets\boldsymbol{\tau}^a\left ( i \right ) $, $\tau^{e,i} \gets\boldsymbol{\tau}^e\left ( i \right ) $ \emph{// Extract the trajectory of the i-th agent}
\STATE Calculate $R^{a,i}_t$,$R^{e,i}_t$ and $A^{a,i}_t$,$A^{e,i}_t$ via Eqn.~\ref{Eq:return} and~\ref{Eq:advantage}. 
\STATE Update $V^a_i$ and $V^e_i$ via Eqn.~\ref{Eq:valueloss}.
\STATE Update $\pi^i$ via Eqn.~\ref{Eq:policyloss}.
\ENDFOR
\ENDFOR
\end{algorithmic}
\end{algorithm}

\subsection{Hybrid Expert-Agent Policy Optimization Module}
\label{Sec: Hybrid policy optimization mechanism}

In this section, we propose the HPO module, a fully decentralized MARL training framework that adaptively combines expert demonstrations from Section~\ref{Sec: Expert Demonstration Mechanism} with autonomous exploration samples, thus accelerating policy convergence and enhancing agent performance.

The HPO module adopts a fully decentralized framework, where the training and execution of each agent are independent to ensure scalability. Specifically, as described in Section~\ref{Sec: Expert Demonstration Mechanism}, the SED module generates the expert trajectory set $\boldsymbol{\tau}^e$ and the agent trajectory set $\boldsymbol{\tau}^a$. For each agent $i$, we denote its own trajectories by $\tau^{e,i}\in \boldsymbol{\tau}^e$ and $\tau^{a,i}\in\boldsymbol{\tau}^a$, where the superscripts $a$ and $e$ indicate trajectories derived from agent-environment interactions and expert instruction executions. To perform policy evaluation and subsequent policy fusion, we define the agent and expert value functions as follows:
\begin{equation}
V^x_i(o^{x,i}_t) = \mathbb{E}_{\tau^{x,i}} \left[ \sum_{k=t}^{\infty } \gamma^{k-t} r^{x,i}_k \right],
\end{equation}
where $x \in \left \{ a, e \right \} $, $o^{x,i}_t$ and $r^{x,i}_k$ are the observation and reward at time step $k$ in trajectory $\tau^{x,i}$. Subsequently, we compute the finite-horizon bootstrapped return for each trajectory using the agent and expert value functions, $V^a_i$ and $V^e_i$, respectively:
\begin{equation}
    \label{Eq:return}
    R^{x,i}_t = \sum_{k=0}^{T^x-t-1} \gamma ^k r^{x,i}_{t+k}+\gamma^{T^x-t} V^x_i\left ( o^{x,i}_{T^x}  \right )
\end{equation}
where $r^{x,i}_{t+k}$ and $o^{x,i}_{T^x}$ are the reward and observation in trajectory $\tau^{x,i}$, and $T^x$ represents the truncation length of the trajectory $\tau^{x,i}$. The agent value function $V^a_i$ and the expert value function $V^e_i$ are optimized by minimizing the mean squared error:
\begin{equation}
    \label{Eq:valueloss}
    \mathcal{L}_{\mathrm{value}}(V^x_i) = \mathbb{E}_{\tau^{x,i}}  \left[ \left(V^x_i(o^{x,i}_t) - R^{x,i}_t \right)^2\right].
\end{equation}

Relying solely on expert demonstration samples often results in limited sample diversity, which constrains the agent’s policy generalization to unknown states. Furthermore, in partially observable environments, LLMs are unable to access complete state information, which may lead to suboptimal expert demonstrations. To address these limitations, we propose a hybrid policy optimization mechanism that integrates both expert demonstration samples and agent autonomous exploration samples. This approach enables the agent to further optimize its policy beyond what can be achieved by learning from expert knowledge alone. Specifically, we first compute the agent and expert advantages for the respective trajectories as follows:
\begin{equation}
    \label{Eq:advantage}
    A^{x,i}_t = R^{x,i}_t - V^x_i(o^{x,i}_t).
\end{equation}
We then adopt the clipped surrogate objective \cite{schulman2017proximal} to construct the agent and expert policy losses using the agent and expert advantages, respectively:
\begin{equation}
    \mathcal{L}^x (\pi^i) = \mathbb{E}_{\tau^{x,i}}\left[ \min(\omega ^{x,i}_t , \mathrm{clip}(\omega ^{x,i}_t, 1 - \epsilon , 1 + \epsilon  ))A^{x,i}_t\right],
\end{equation}
where $\omega^{x,i}_t = \pi^i(u^{x,i}_t|o^{x,i}_t) / \pi^{\mathrm{old},i}(u^{x,i}_t|o^{x,i}_t)$ denotes the importance sampling ratio, $\pi^{\mathrm{old},i}$ denotes the policy of agent $i$ from the previous iteration, and $\epsilon \in (0, 1)$ represents the clipping coefficient. Consequently, we define the hybrid policy loss function $\mathcal{L}_\mathrm{mix}$ as follows:
\begin{equation}
\label{Pimaxloss}
\mathcal{L}_\mathrm{mix} (\pi^i) = \alpha\mathcal{L}^a (\pi^i) + (1-\alpha) \mathcal{L}^e (\pi^i),
\end{equation}
where $\alpha = \exp\left(-k/K \cdot \mathrm{D}_\mathrm{DTW}(\tau^{a,i}, \tau^{e,i})\right)$ dynamically adjusts the weighting between the agent and expert policy losses, with $k$ and $K$ denoting the current and total training epochs, respectively, and $\mathrm{D}_\mathrm{DTW}(\tau^{a,i}, \tau^{e,i})$ denotes the dynamic time warping (DTW) distance, which measures the similarity between agent and expert trajectories via nonlinear temporal alignment. To promote autonomous exploration, we incorporate a maximum entropy regularization term into the hybird policy loss function, which is defined as follows:
\begin{equation}
    \label{Eq:policyloss}
   \mathcal{L} (\pi^i)=\mathcal{L}_\mathrm{mix} (\pi^i) + \beta \mathbb{E}_{\tau^{a,i}}[ \mathcal{H}(\pi^i(\cdot|o^{a,i}_t)) ],
\end{equation}
where $\beta \in \left(0, 1\right]$ denotes the entropy regularization coefficient, and $\mathcal{H}$ denotes the entropy of the policy distribution.

As shown in Figure~\ref{fig:2}, during the initial training phase, the agent lacks sufficient experience, which leads to a substantial discrepancy between its policy and the expert demonstrations. The resulting large DTW distance induces a low hybrid weight $\alpha$ that prioritizes expert-guided policy optimization. As training progresses and the agent’s policy converges towards the expert’s behavior, the DTW distance decreases, leading to a gradual increase in $\alpha$ and the optimization increasingly emphasizes experiences gained through agent's autonomous exploration. In parallel, the SED module will periodically leverage environmental feedback to refine and provide updated demonstrations, mitigating the non-stationarity resulting from current policy updates. With each refinement, the difference between the previous and new demonstrations enlarges the DTW distance, which reduces $\alpha$ and guides the agent to optimize its policy to follow the new demonstrations.This iterative mechanism, integrating both expert guidance and self-exploration, enables the agent to continuously improve its policy, thereby achieving efficient learning and improved generalization across complex tasks.

Algorithm~\ref{Alg:Training Procedure} outlines the training procedure of RELED, where $q$ denotes the sampling interval for expert demonstration. The training process begins with the initialization of the agent policy $\pi^i$, agent value function $V^{a,i}$, and expert value function $V^{e,i}$ for each agent $i$ (line 1). Subsequently, the LLM $\Pi$ is loaded, and the initial prompt $d_{\mathrm{init}}$ is set (line 2). At each training epoch, the agent trajectory set $\boldsymbol{\tau}^a$ is collected by executing the current joint policy in the environment (line 4). To improve the stability of policy updates, the SED module adopts a sampling interval $q$, allowing each agent to utilize the same expert demonstration over multiple epochs (lines 6-17). During each demonstration generation, the agent set $\mathcal{A}$ is randomly partitioned into $m$ disjoint subsets (line 7). For each subset $\mathcal{A}_j$, expert trajectories $\boldsymbol{\tau}^{e,j}$ are obtained by applying LLM-generated instructions to agents in $\mathcal{A}_j$, while the remaining agents follow their own policy decisions (line 11). The reward-instruction pair set $\mathcal{R}^j$ and the policy divergence-instruction pair set $\mathcal{K}^j$ are calculated to quantitatively assess the impact of the demonstration on environment stationarity (line 12), and these results are integrated into the feedback prompt for LLM refinement (line 13). Subsequently, the observation information from the expert trajectory set $\boldsymbol{\tau}^e$ is incorporated into the feedback prompts to provide the LLM with detailed context on  expert instruction executions (line 16). In the HPO module, each agent extracts its own agent trajectory and expert trajectory for policy evaluation and improvement (line 20). Bootstrapped returns $R^{a,i}_t$, $R^{e,i}_t$ and advantages $A^{a,i}_t$, $A^{e,i}_t$ are calculated (line 21), serving as inputs to update agent and expert value functions via mean squared error loss (line 22), and to optimize policy $\pi^i$ using the hybrid loss with entropy regularization (line 23).

\section{Implementation}
This section describes the experimental setups and implementation details of RELED.
\label{Sec: Implementation}

\subsection{Environment Setups}

\textit{1) Simulation:} We evaluate RELED in multi-agent navigation tasks in realistic urban traffic environments constructed from OpenStreetMap (OSM) data \cite{OpenStreetMap} and simulated in SUMO \cite{SUMO2018}. OSM data are converted to SUMO via a standard netconvert pipeline, including geometry simplification, ramp inference, junction merging, standardizing intersections and signals while preserving lane counts, one-way attributes, and speed limits.

Each agent controls a single vehicle navigating from an assigned origin to a designated destination. For fair comparison, origin–destination pairs are sampled once at random from a predefined set and then fixed identically for all methods and runs. A single environment interaction step corresponds to one agent perform an action. The simulator advances in discrete steps last up to 2400 simulator steps in total, terminating early if all agents reach their destinations.

\begin{figure}
    \centering
    \includegraphics[width=0.9\linewidth]{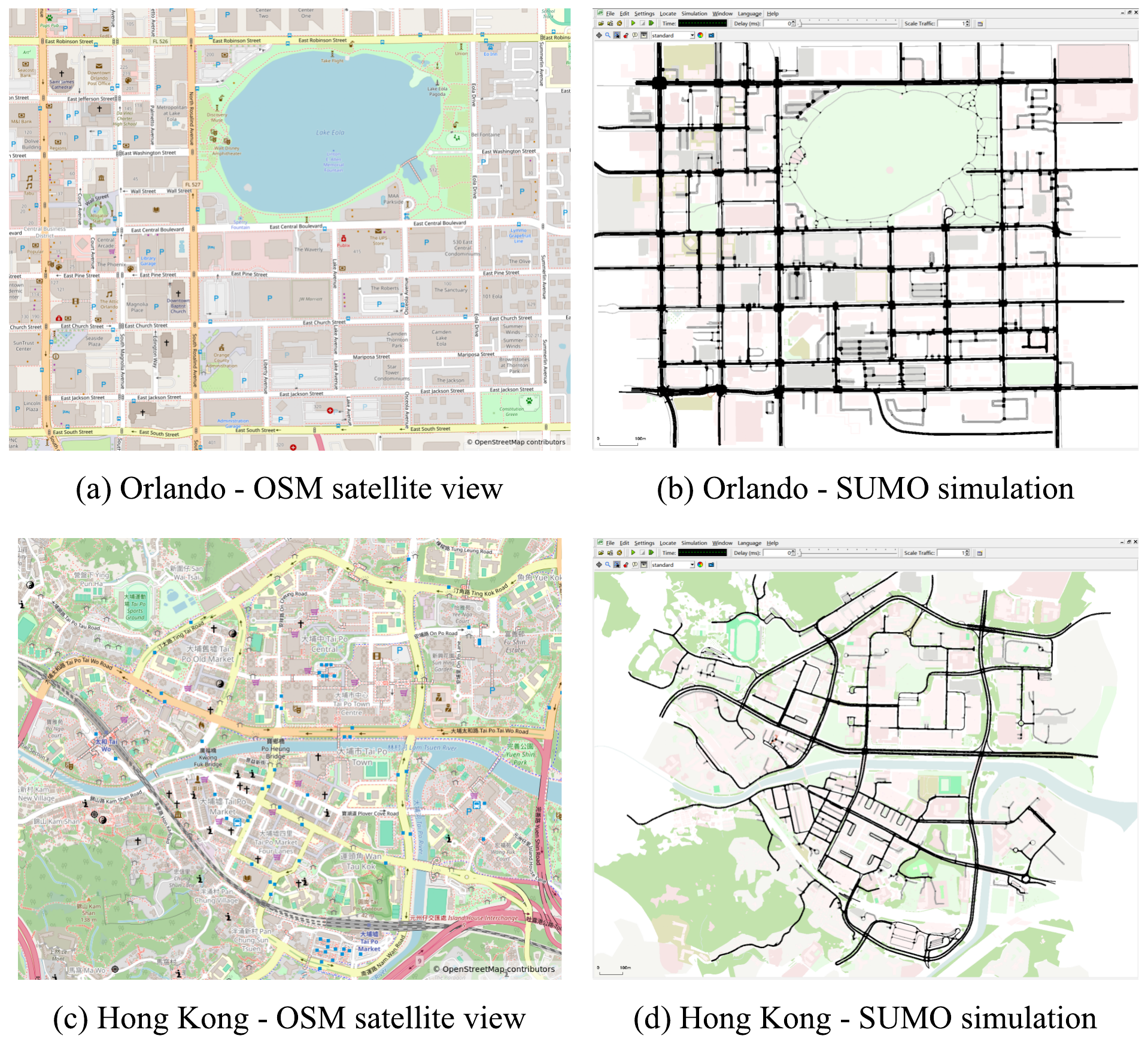}
      \vspace{-2ex}
    \caption{Experimental scenarios.}
    \label{fig:0_maps}
\end{figure}

For each map, we consider two background traffic regimes to emulate different demand levels: (i) moderate flow, with 90 background vehicles distributed according to edge lengths; and (ii) congested flow, with 190 background vehicles and increased injection frequency on major arterials. Background vehicles has randomness in route assignment, departure times, and lane selection, and their origin–destination pairs are sampled from the same predefined set as the agents. The experiments span two representative urban networks:

\begin{itemize}
    \item \textbf{Orlando:} A canonical grid with uniform intersections (Figure~\ref{fig:0_maps}(a,b)), featuring 115 regulated intersections and 269 edges across a 1.99 km² area. This environment primarily consists of  four-way junctions with consistent lane configurations.
    \item \textbf{Hong Kong:} An urban road network (Figure~\ref{fig:0_maps}(c,d)) spanning 3.87 square kilometers, comprising 144 regulated intersections and 239 road segments, characterized by roads with extended edge lengths and a high prevalence of one-way traffic routes.
\end{itemize}

\begin{table}[t]
\centering
\caption{Action space and observation space.}
\label{tab:action_obs_space}
\resizebox{0.45\textwidth}{!}{%
\setlength{\tabcolsep}{3pt}
\renewcommand{\arraystretch}{0.75}
\begin{tabular}{c|l}
\toprule
\multicolumn{2}{c}{\textbf{Action Space (Discrete, shape: $1$)}} \\
\midrule
\textbf{Index} & \textbf{Note} \\
\midrule
$0$ & Select the $i$-th available edge ($i<m_{out}^*$) \\
\midrule
\multicolumn{2}{c}{\textbf{Observation Space (Shape: $2m_{out}+2$)}} \\
\midrule
\textbf{Index} & \textbf{Note} \\
\midrule
$[0$ & Current junction id \\
$[1]$ & Destination junction id \\
$[2 + 2i]$ & Score of the $i$-th available edge ($i=0..m_{out}-1$) \\
$[2 + 2i + 1]$ & End junction id of the $i$-th edge ($i=0..m_{out}-1$) \\
\bottomrule
\end{tabular}
}
\begin{tablenotes}
\item[$*$] $m_{out}$ is the maximum number of outgoing edges at any junction.
\end{tablenotes}
\end{table}


\textit{2) Hyperparameters: } Each independent run consisted of $5 \times 10^{2}$ training epochs and each epoch comprising $10^{3}$ environment interaction steps, accumulating a total of $5 \times 10^{5}$ interaction steps. Unless otherwise specified, all experiments were conducted with 10 agents operating in the environment. All algorithms are configured with identical optimizer parameters, with the learning rate $= 3 \times 10^{-4}$ and The discount factor $\gamma = 0.99$. Both policy networks and value estimation networks across all algorithms employed the same two-layer neural network structure with 128 neurons per layer. For LLM demonstration sampling, we employed a periodic update strategy with interval $q=10$ epochs. For RELED, the agent subset partitioning parameter $m = 2$.

\textit{3) Reward Function: }
The reward at interaction step k is designed to encourage minimal travel time and progress toward the destination:
\begin{equation}
R_k = -(t_k - t_{k-1}) + \omega_d(d_{k-1} - d_k),
\end{equation}
where $R_k$ is the reward at step $k$; $t_k$ and $t_{k-1}$ are the current and previous simulation times; $\omega_d$ is set to $1$ in our experiments; and $d_k$ and $d_{k-1}$ are the current and previous Euclidean distances to the destination. The first term penalizes elapsed time, and the second term rewards reductions in distance to the destination—when the distance decreases, the agent receives a positive reward proportional to that decrease. Upon reaching the destination, the agent receives an additional terminal reward $R_{\text{term}} = 0.1 \cdot T_{\max}$, where $T_{\max}$ is the maximum number of simulation steps.

\textit{4) Observation Space: }
Agents operate with partially observation space, with visibility limited to all outgoing edges of the junction they are approaching. The observation space of a single agent contains the current junction ID, the destination junction ID, and for each potential outgoing edge, its score and the end junction ID. The edge score ranges from 0 to 1, calculated as the ratio between the free flow travel time (length/max speed) and the estimated travel time. Detailed environment specifications are summarized in Table 1.

\textit{5) Action Space: }
The action space is a discrete selection of available road segments at each junction. When an agent reaches an intersection, it must choose one outgoing edge from $m_{out}$ options. For each map, $m_{out}$ is fixed and set to the maximum number of outgoing roads at any junction in that map. For junctions with fewer than $m_{out}$ edges, the excess dimensions are masked with a $-1$. For the Orlando and Hong Kong maps used in our experiments, $m_{out} = 4$.

\begin{table}[t]
\centering
\caption{Predefined navigation interfaces provided for LLM}
\label{tab:api}
\resizebox{0.45\textwidth}{!}{%
\setlength{\tabcolsep}{3pt}%
\renewcommand{\arraystretch}{0.9}%
\begin{tabular}{l|l}
\toprule
\multirow{2}{*}{\textbf{Function}} & \multirow{2}{*}{\textbf{Description}} \\
 & \\
\midrule
move\_to\_by\_shortest\_path(node\_id) $\rightarrow$ bool & Move to target node via shortest distance path \\
\midrule
move\_to\_by\_shortest\_time(node\_id) $\rightarrow$ bool & Move to target node via shortest time path \\
\midrule
get\_origin() $\rightarrow$ int & Return the current agent's origin node ID \\
\midrule
get\_destination() $\rightarrow$ int & Return the current agent's destination node ID \\
\midrule
get\_shortest\_dist(target\_node\_id) $\rightarrow$ float & Calculate shortest path distance to target node \\
\midrule
get\_shortest\_time(target\_node\_id) $\rightarrow$ float & Calculate shortest time to target node \\
\midrule
get\_nearest\_node(x, y) $\rightarrow$ int & Find node ID closest to coordinates (x, y) \\
\midrule
get\_node\_coord(node\_id) $\rightarrow$ tuple[float, float] & Get coordinates of specified node \\
\bottomrule
\end{tabular}
}%
\end{table}

\subsection{Baselines}
To evaluate the performance of RELED, we compare it with the following MARL methods:

\begin{itemize}
\item \textbf{IPPO}~\cite{de2020independent}:  An extension of PPO where each agent is trained independently with its own policy and value networks, using on-policy data.

\item \textbf{MAPPO}~\cite{yu2022surprising}: A CTDE method that trains decentralized policies together with a centralized value function that has access to global information during training.

\item \textbf{QMIX}~\cite{rashid2020monotonic}: A value factorization approach that combines individual agent value functions through a monotonic mixing network to form a joint action-value function for decentralized control.

\end{itemize}

\subsection{RELED Prompt Design}
We implement a prompting architecture to generate high-quality LLM-based expert demonstrations for the RELED framework. We employ GPT-3.5 Turbo as our foundational model and structure our prompting approach in a two-phase methodology. Detailed prompt templates and example responses can be found in Appendix A. 

\textit{1) Initial Prompt: }
The initial prompt defines the task and outlines the context of the environment. The model receives the network topology (node coordinates and a directed adjacency list), each agent’s origin–destination pair, and its departure time. We also provide a small set of navigation utility interfaces (Table~\ref{tab:api}), which are thin wrappers around SUMO's built-in path planning and information routines. Finally, we specify output constraints with instructions to generate exeutable Python code format. In practice, the above prompt serves as the system prompt for the model. To maintain computational efficiency, the context window is limited to 10,000 tokens; earlier historical information is discarded as needed, while the system prompt is preserved throughout the interaction process. In addition, the initial prompt contains a simple command to trigger the instruction.

\textit{2) Feedback Prompt:}
Every $q=10$ epochs, we compile a summary of recent agent exploration and expert-generated plans and deliver it to the LLM as a structured feedback prompt. The summary includes two diagnostics: \textit{(i) reward volatility index (RVI)}: As defined in Eqn.~\ref{Eq:reward volatility index}, RVI identifies groups whose substituted expert policies produce the largest single-step reward decreases for agents running the current DRL policy, together with the associated states and triggering instructions; \textit{(ii) the policy divergence index (PDI)}: As defined in Eqn.~\ref{Eq:policy divergence index}, PDI computed as the KL divergence between expert and learned policies at matched decision states, together with the associated states and triggering instructions. Using the combined information, the LLM revises the expert plans to mitigate cross-agent interference and reduce policy mismatch.

\section{Evaluation}
\label{Sec: Evaluation}

\begin{figure*}
    \centering
    \includegraphics[width=1\linewidth]{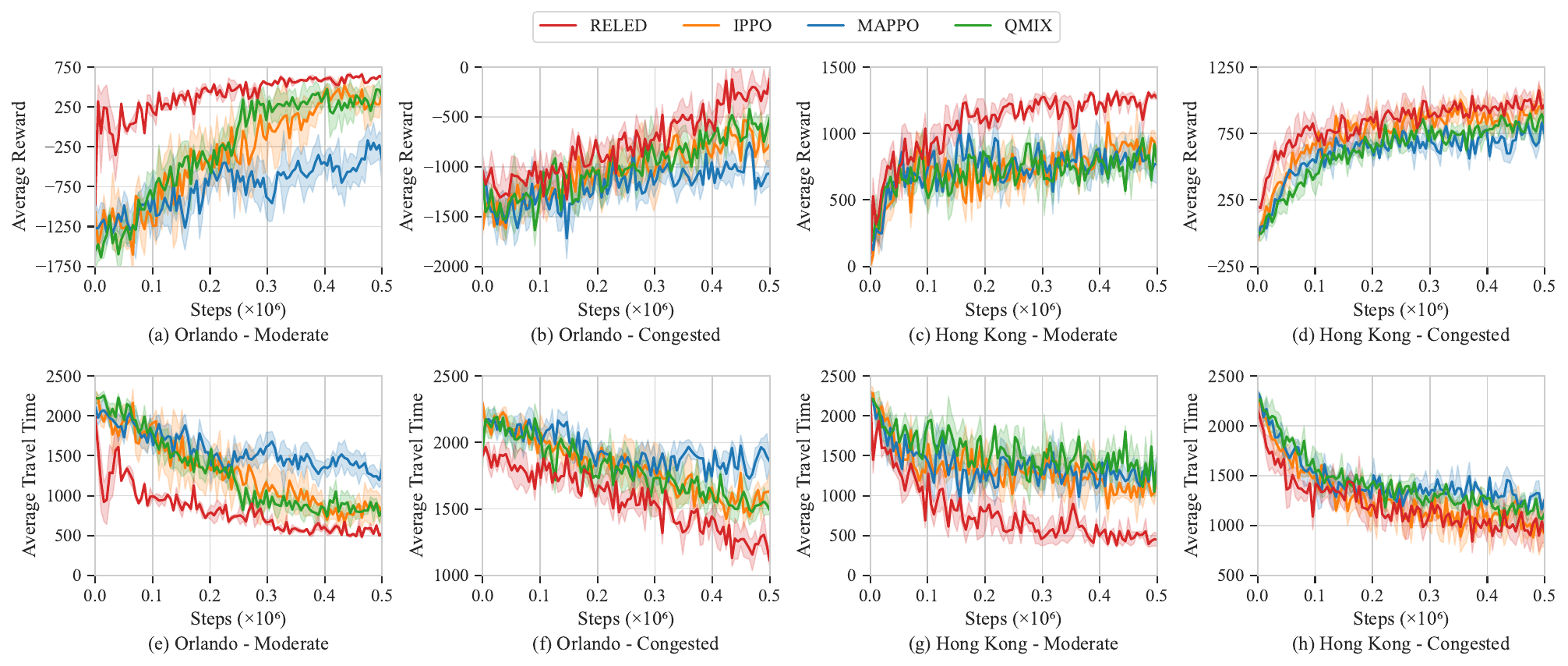}
    \caption{Sample efficiency across cities and traffic regimes. (a–d) Average episode reward vs. interaction steps under Moderate and Congested settings. (e–h) Average travel time (measured in SUMO simulator time step) vs. steps. Shaded regions denote SD over 3 independent runs.}
    \label{fig:sample_ef}
\end{figure*}
In this section, we conduct extensive experiments to evaluate the effectiveness of our proposed method RELED. 

\subsection{Comparative Results}
This section compare RELED with state-of-the-art MARL methods on sample efficiency and time efficiency.

\textit{1) Sample Efficiency: }
Figure~\ref{fig:sample_ef} demonstrates that RELED consistently outperforms all baseline approaches in both sample efficiency and final performance across all four test scenarios. RELED achieves higher average episode rewards with fewer training steps. In the moderate traffic conditions of (a) Orlando and (c) Hong Kong, RELED rapidly reaches strong performance levels and continues to improve steadily. In the more challenging congested environments of (b) Orlando and (d) Hong Kong, RELED maintains consistently higher average rewards per interaction step compared to all baseline methods. The complementary metrics average travel time presented in Figure~\ref{fig:sample_ef} (e)-(h) provide further evidence of RELED's superior performance. Average travel time directly reflects agent effectiveness in traffic management scenarios as a pure optimization objective, unaffected by reward shaping components. Time unit here is SUMO simulator time step, which is the standard temporal measurement in the SUMO traffic simulation environment. RELED reduces travel times more rapidly and converges to lower values than the baselines.  This highlights RELED's enhanced ability to efficiently translate sample experiences into effective policy improvements across diverse urban networks and traffic conditions.

\begin{figure}
    \centering
    \includegraphics[width=1\linewidth]{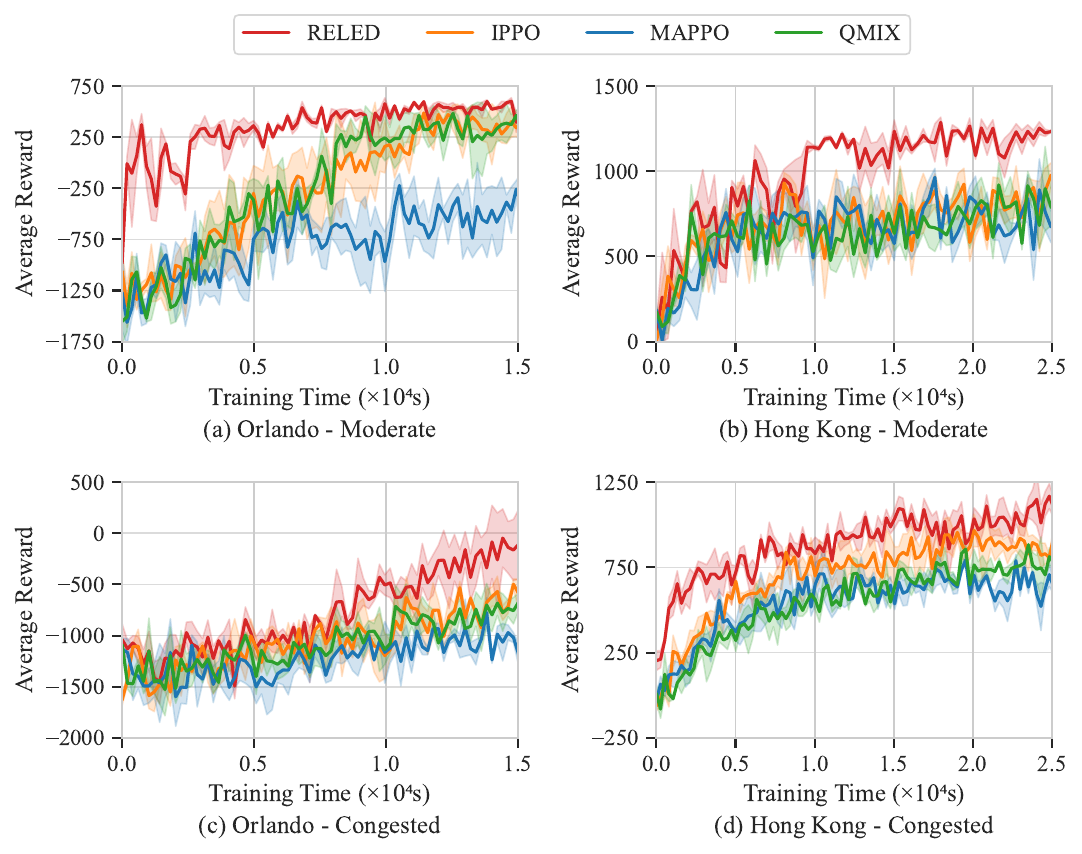}
    \caption{Time efficiency in wall-clock training. Average episode reward over elapsed time for Orlando and Hong Kong (Moderate and Congested). Shaded regions indicate SD across 3 independent runs.}
    \label{fig:time_ef}
\end{figure}

\textit{2) Time Efficiency: }
Figure~\ref{fig:time_ef} illustrates the progression of average episode reward over training time. Despite the additional computational overhead introduced by LLM inference, RELED remains competitive in time efficiency. For Orlando – moderate, even with the maximum historical context, the average time for a single refinement inference is only 7.35 ± 4.86 seconds, which is small relative to the overall training duration. More detailed analysis of model inference time is presented in Section~\ref{Sec: SED}. Across both urban environments and traffic conditions, RELED achieves higher average rewards earlier than all baselines, establishing an early performance advantage that is maintained throughout the training process. These results shows that the incremental inference cost is more than offset by the resulting more informative updates, yielding greater performance gains per unit time.

\subsection{Sample Analysis of HPO Module}
This section presents empirical evidence that our method effectively aligns agents with expert behavior and transitions learning from imitation to self-driven optimization in HPO module.

\begin{figure}
    \centering
    \includegraphics[width=1\linewidth]{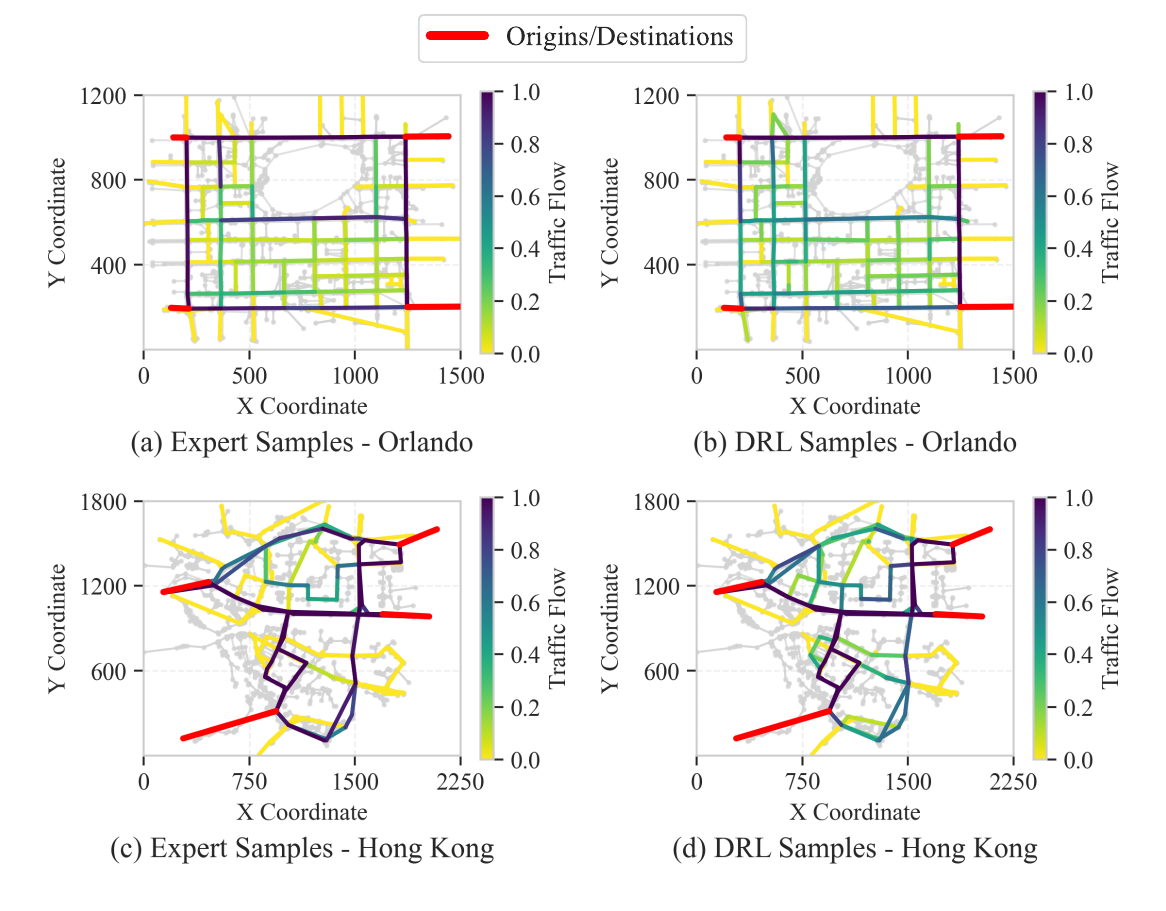}
    \caption{Qualitative trajectory visualization comparing expert demonstrations (left) and agent-generated trajectories (right) in  (a, b) Orlando - Moderate and (c, d) Hong Kong - Moderate.}
    \label{fig:visual}
\end{figure}

\textit{1) Trajectories Visualization:}
Figure~\ref{fig:visual} visualizes expert demonstrations and agent-generated trajectories in Orlando - Moderate and Hong Kong - Moderate. The visualization presents samples from the final training phase, showing the last 10 episodes for both expert and agent trajectories.The data is log-transformed and normalized to [0,1] to better represent the right-skewed distribution of traffic flow. In both environments, expert demonstrations reveal concentrated traffic flow along key corridor chains connecting origins and destinations. DRL agents successfully replicate these primary corridors while maintaining comparable flow intensities on critical segments, demonstrating effective policy transfer. Differences primarily emerge in secondary connections near intersections, where the agents redistribute some traffic flow. This reflects the inherent exploration-exploitation trade-off in reinforcement learning. Agents leverage expert demonstrations as a structural foundation while conducting exploration on peripheral edges to discover potentially improved routing decisions.

\begin{figure}
    \centering
    \includegraphics[width=1\linewidth]{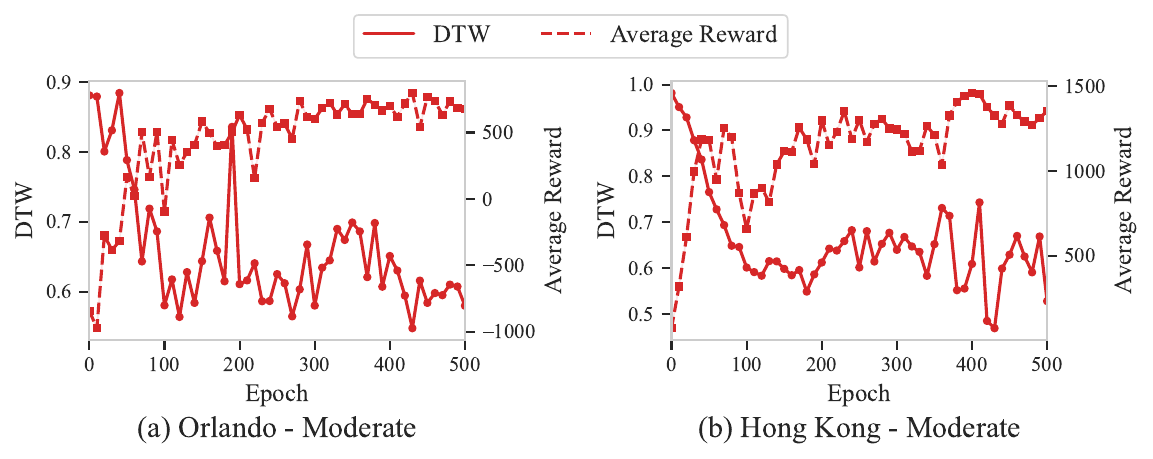}
    \caption{DTW dynamics with training. DTW distance between expert and agent trajectories (left axis) alongside average episode reward (right axis) during training on (a) Orlando - Moderate and (b) Hong Kong - Moderate.}
    \label{fig:dtw}
\end{figure}

\textit{2) Policy Alignment: }
To quantify this alignment, Figure~\ref{fig:dtw} reports the DTW distance between expert and agent trajectories across training, together with the average episode reward. DTW measures the similarity between temporal sequences by finding the optimal nonlinear alignment between them. Lower DTW denotes closer adherence to the expert policy. In both environment, DTW drops quickly in the early epochs and stays relatively low thereafter, while reward rises. In RELED, a lower DTW leads to a reduced weighting of expert samples in policy updates through the coefficient $\alpha$, as expressed in Eqn.~\ref{Pimaxloss}. The result shows that the agents gradually shifts learning toward self-generated experience, continuously improving performance even after achieving close alignment with expert behavior.

\subsection{Demonstration Analysis of SED Module}
\label{Sec: SED}
In this section, we evaluate LLM-generated demonstrations across various refinement phases, model choices, and prompt designs. We also identify how the SED module contributes to high-quality demonstrations through its structured interfaces and iterative feedback mechanisms.

\textit{1) LLM Model Choice:}
Table~\ref{tab:llm_demo} presents our evaluation of LLM-generated demonstrations across three refinement phases in Orlando - Moderate. For each phase, 100 expert trajectories (10 prompts × 10 agents) are collected. We evaluated four widely-used LLMs of varying parameter sizes for their impact on the quality of expert demonstrations. GPT-4 Turbo achieves the highest rewards (752.84 ± 65.32 with full setting at 50 iterative refinements) but requires substantially longer inference times (19.55 ± 1.70 seconds). GPT-3.5 Turbo and Llama-3.1-70b show competitive performance (684.76 ± 57.34 and 708.46 ± 47.28 rewards) with faster inference (7.35 ± 4.86 and 6.86 ± 1.35 seconds), offering excellent balance between quality and computational efficiency. In our experiments, we selected GPT-3.5 Turbo for this balanced performance profile. Overall, larger models tend to deliver higher rewards and execution rates on average, whereas smaller models like Llama-3.1-8B are attractive  with strict computational constraints (e.g., 4.64 ± 0.60 s at 50 refinements).

\textit{2) Impact of Prompt Design:}
The ablation study of prompt design in Table~\ref{tab:llm_demo} highlights how different components impact the quality of expert demonstrations. In the absence of refinement feedback (Refn.), DTW difference decreases substantially. For example, with GPT-3.5 Turbo (50 refinements), the DTW difference drops to 836.66 ± 206.82 (vs. 1040.50 ± 174.15 with both components). This pronounced reduction in DTW indicates that the feedback mechanism, by iteratively refining model outputs, enhances the consistency between expert demonstrations and reinforcement learning samples. Providing LLMs with predefined navigation interfaces (IFs) generally improves performance, compared to letting LLM generate complete node-connected routes. This advantage is more pronounced in higher refinement iterations where there is more context. At 50 refinements, without IFs, execution rates decline dramatically while rewards decrease substantially — GPT-3.5 Turbo's execution rate falls from 85\% to 67.00\% and rewards drop from 684.76 ± 57.34 to 428.92 ± 62.16, with inference time also increasing significantly from 7.35 ± 4.86 to 12.46 ± 1.40.

\begin{table*}[t]
\centering
\caption{LLM-generated demostrations under different settings and refinement iterations}
\label{tab:llm_demo}
\resizebox{\textwidth}{!}{%
\setlength{\tabcolsep}{3pt}%
\renewcommand{\arraystretch}{0.9}%
\begin{tabular}{l|cc|ccc|cccc|cccc}
\toprule
\multirow{2}{*}{Model} & \multirow{2}{*}{IFs} & \multirow{2}{*}{Refn.} & \multicolumn{3}{c|}{Initial} & \multicolumn{4}{c|}{25 Refinements} & \multicolumn{4}{c}{50 Refinements} \\
 & & & Exec. (\%) ↑ & Reward ↑ & Time ↓ & Exec. (\%) ↑ & Reward ↑ & Time ↓ & DTW diff. ↑ & Exec. (\%) ↑ & Reward ↑ & Time ↓ & DTW diff. ↑ \\
\midrule
\multirow{3}{*}{GPT-4 Turbo} & $\checkmark$ & $\checkmark$ & 92.00$\pm$16.00 & 687.92$\pm$78.01 & 14.36$\pm$2.49 & 77.00$\pm$10.05 & 724.63$\pm$72.15 & 19.83$\pm$2.43 & 1266.75$\pm$324.20 & 80.00$\pm$8.94 & 752.84$\pm$65.32 & 19.55$\pm$1.70 & 980.39$\pm$220.59 \\
 & $\checkmark$ & & 74.00$\pm$36.66 & 643.10$\pm$42.68 & 13.28$\pm$2.01 & 79.00$\pm$8.31 & 685.27$\pm$38.42 & 19.55$\pm$1.28 & 928.82$\pm$156.94 & 74.00$\pm$20.10 & 710.38$\pm$35.76 & 16.58$\pm$0.98 & 949.72$\pm$197.49 \\
 & & $\checkmark$ & 50.00$\pm$36.06 & 447.96$\pm$80.57 & 35.55$\pm$4.87 & 34.00$\pm$39.04 & 495.74$\pm$73.26 & 53.61$\pm$9.93 & 1066.75$\pm$222.53 & 52.00$\pm$24.41 & 543.82$\pm$67.38 & 52.54$\pm$7.30 & 943.87$\pm$197.46 \\
\midrule
\multirow{3}{*}{GPT-3.5 Turbo} & $\checkmark$ & $\checkmark$ & 92.00$\pm$4.00 & 608.42$\pm$68.53 & 5.45$\pm$1.13 & 92.00$\pm$16.00 & 645.32$\pm$62.46 & 7.67$\pm$4.57 & 1012.38$\pm$187.65 & 85.00$\pm$6.71 & 684.76$\pm$57.34 & 7.35$\pm$4.86 & 1040.50$\pm$174.15 \\
 & $\checkmark$ & & 86.00$\pm$22.00 & 556.74$\pm$58.32 & 4.30$\pm$1.89 & 87.00$\pm$15.52 & 584.38$\pm$52.76 & 7.01$\pm$4.56 & 817.46$\pm$142.38 & 80.00$\pm$7.75 & 615.82$\pm$47.28 & 6.88$\pm$0.90 & 836.66$\pm$206.82 \\
 & & $\checkmark$ & 54.00$\pm$18.00 & 584.86$\pm$72.48 & 6.51$\pm$1.63 & 45.00$\pm$15.00 & 386.35$\pm$68.24 & 12.60$\pm$5.01 & 812.54$\pm$174.28 & 67.00$\pm$11.87 & 428.92$\pm$62.16 & 12.46$\pm$1.40 & 1222.51$\pm$209.50 \\
\midrule
\multirow{3}{*}{Llama-3.1-70b} & $\checkmark$ & $\checkmark$ & 83.00$\pm$14.18 & 625.38$\pm$57.46 & 5.33$\pm$0.41 & 86.00$\pm$8.00 & 668.75$\pm$52.38 & 6.46$\pm$0.58 & 1112.64$\pm$212.38 & 74.00$\pm$6.63 & 708.46$\pm$47.28 & 6.86$\pm$1.35 & 1017.66$\pm$117.68 \\
 & $\checkmark$ & & 86.00$\pm$8.00 & 586.42$\pm$48.26 & 6.02$\pm$0.58 & 91.00$\pm$9.43 & 617.38$\pm$42.76 & 4.47$\pm$0.38 & 863.54$\pm$152.63 & 70.00$\pm$14.83 & 658.74$\pm$38.42 & 4.78$\pm$0.93 & 941.86$\pm$104.88 \\
 & & $\checkmark$ & 59.00$\pm$17.58 & 684.62$\pm$53.84 & 6.60$\pm$1.66 & 64.00$\pm$36.66 & 527.35$\pm$47.62 & 9.76$\pm$1.43 & 916.78$\pm$163.42 & 58.00$\pm$20.88 & 568.42$\pm$42.16 & 14.32$\pm$1.23 & 1098.65$\pm$220.97 \\
\midrule
\multirow{3}{*}{Llama-3.1-8b} & $\checkmark$ & $\checkmark$ & 68.00$\pm$8.93 & 613.98$\pm$48.55 & 4.79$\pm$0.31 & 68.00$\pm$16.61 & 591.46$\pm$56.19 & 4.89$\pm$0.40 & 762.59$\pm$294.72 & 69.00$\pm$25.08 & 595.67$\pm$43.05 & 4.64$\pm$0.60 & 846.96$\pm$167.29 \\
 & $\checkmark$ & & 79.00$\pm$8.31 & 557.56$\pm$89.99 & 4.39$\pm$0.22 & 63.00$\pm$24.52 & 512.08$\pm$35.09 & 3.59$\pm$0.21 & 781.50$\pm$100.24 & 69.00$\pm$23.00 & 526.66$\pm$87.94 & 3.70$\pm$0.81 & 710.23$\pm$198.95 \\
 & & $\checkmark$ & 47.00$\pm$7.81 & 519.18$\pm$38.02 & 2.79$\pm$0.60 & 60.00$\pm$48.99 & 607.84$\pm$64.84 & 11.49$\pm$4.74 & 749.58$\pm$163.29 & 42.00$\pm$20.40 & 510.51$\pm$94.75 & 10.06$\pm$5.59 & 714.42$\pm$153.39 \\
\bottomrule
\end{tabular}
}%
\begin{tablenotes}
\item[$*$] We report execution rate (Exec.), average reward of the demonstration trajectory (Reward), LLM inference time (Time), DTW differences relative to initial trajectories (DTW diff.), existence of navigation interface (IFs) and interactive refinement mechanism (Refn.) as mean ± SD over runs. The ↑ indicates higher value is better, while the ↓ indicates lower value is better.
\end{tablenotes}
\end{table*}

\begin{figure}
    \centering
    \includegraphics[width=1\linewidth]{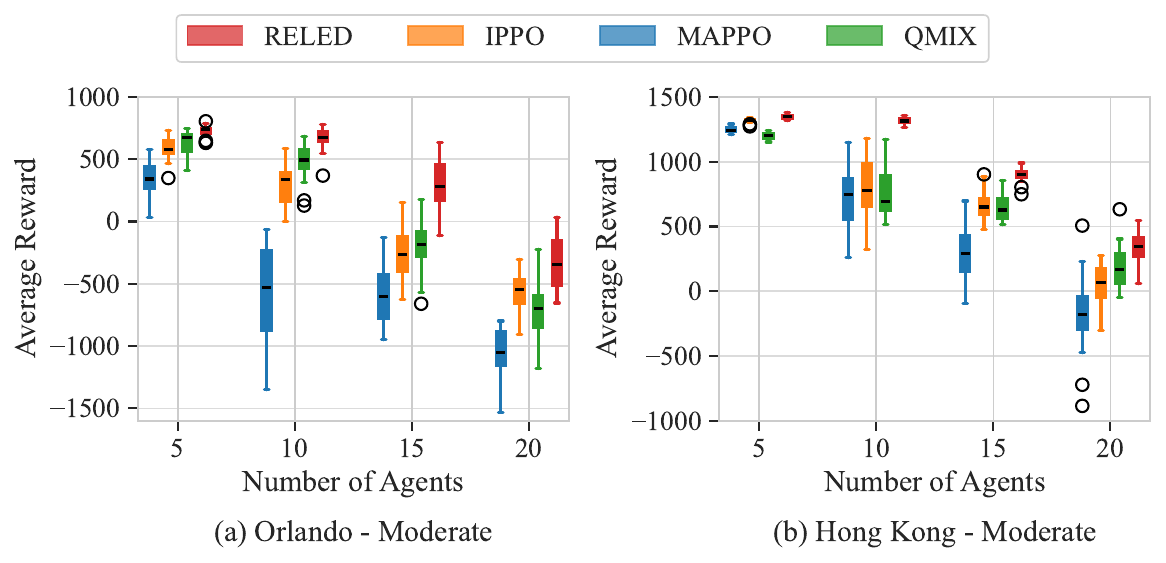}
    \caption{Performance under different numbers of agents, reported by the distribution of average episode rewards over 20 test runs on (a) Orlando - Moderate and (b) Hong Kong - Moderate.}
    \label{fig:scalability}
\end{figure}

\subsection{Scalability}
This section evaluates scalability of RELED. Figure~\ref{fig:scalability} evaluates performance as the number of agents increases on the Orlando - Moderate and Hong Kong - Moderate scenarios. Each box shows the distribution of the average episode reward after 20 test episodes of these models. Across all scales, RELED maintains the highest median reward compared to IPPO, MAPPO, and QMIX. As the population grows from 5 to 20, all methods experience some degradation, but RELED's decline is more gradual. These results indicate that the proposed approach scales more robustly with the number of agents.

\begin{figure}
    \centering
    \includegraphics[width=1\linewidth]{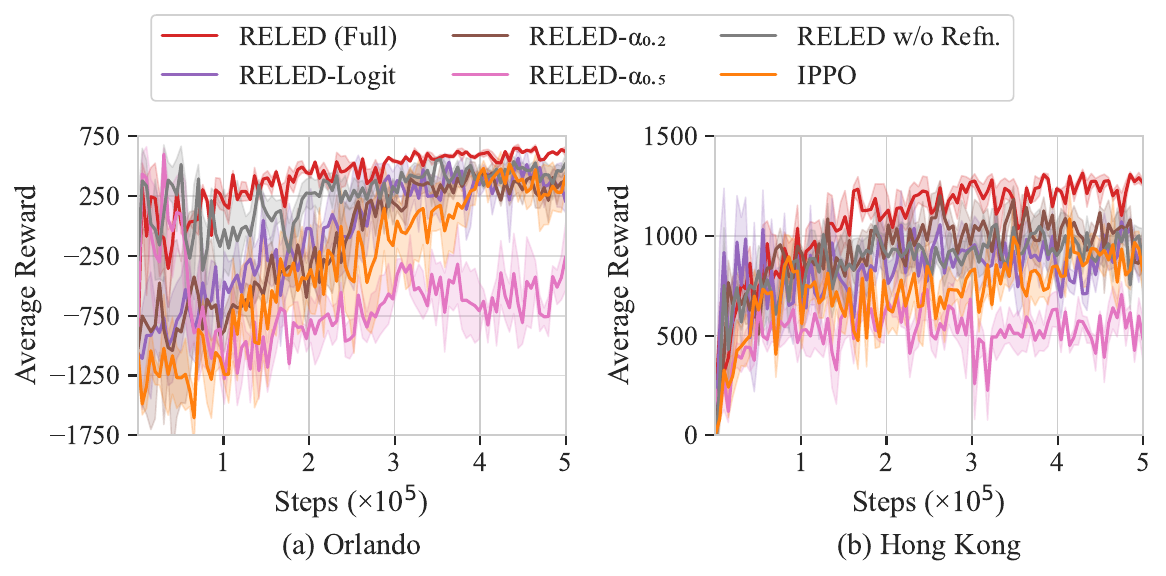}
    \caption{Ablation study on training dynamics comparing average reward over steps for RELED (Full) and alternative variants on (a) Orlando - Moderate and (b) Hong Kong - Moderate; shaded regions indicate SD across 3 runs.}
    \label{fig:ablation}
\end{figure}

\subsection{Ablation Study}
This section evaluates the impact of each module of RELED performance. We compare the following variants:
\textit{(i) RELED (Full)};
\textit{(ii) RELED w/o Refn.}: Implemented without iterative prompt refinement;
\textit{(iii) RELED-$\alpha_{0.2}$};
\textit{(iv) RELED-$\alpha_{0.5}$}: Using fixed expert weights $\alpha=0.2$ and $\alpha=0.5$ instead of dynamic $\alpha$ calculated by DTW distance;
\textit{(v) RELED-Logit}: Replacing LLM-generated demonstrations with a logit-based softmax over shortest-path costs to prioritize lower-cost routes;
and \textit{(vi) IPPO}.

\textit{1) Training Performance: }
Figure \ref{fig:ablation} presents our ablation study comparing these RELED variants in training performance. RELED (Full) achieves the fastest performance gains and the highest episode reward in both environments with reduced variance throughout training. This suggests that iteratively updating expert demostrations can help improve training efficiency. Removing iterative refinement (RELED w/o Refn.) slows late-stage reward improvement. Using fixed expert weights RELED-$\alpha_{0.2}$/RELED-$\alpha_{0.5}$) further reduces sample efficiency; an overly large expert weight ($\alpha=0.5$) leads to insufficient exploration and a sharp late-stage reward drop, especially in (a) Orlando. This validates the advantage of DTW-based adaptive weighting for scheduling the transition from imitation to exploration. RELED-Logit shows higher early sample efficiency but overall lags behind RELED (Full), suggesting that LLM-derived trajectories provide higher-quality, more context-aware guidance than logit-stochastic routing.

\textit{2) Converged Performance: }
Table~\ref{tab:performance_comparison} reports the performance of the convergent model of each method, tested over 20 independent runs. RELED (Full) achieves the highest rewards with minimal travel times in both networks. RELED-Logit attains 43.37\% lower reward and 65.24\% longer travel time than RELED (Full) in Orlando, with similar gaps in Hong Kong. Fixed-weight configurations show severe performance degradation, with $\alpha=0.5$ yielding negative rewards in Orlando (-109.65\%) and substantial drops in Hong Kong (-53.86\%), while $\alpha=0.2$ also degrades performance in both environments (-34.41\% and -23.39\% lower rewards). Removing refinement (RELED w/o Refn.) results in modest performance drops across both environments (-19.79\% and -26.15\% lower rewards). These results quantitatively demonstrate the impact of each RELED component on overall converged performance.

\begin{table}[t]
\centering
\caption{Convergence performance of different variants.}
\label{tab:performance_comparison}
\resizebox{0.45\textwidth}{!}{%
\setlength{\tabcolsep}{3pt}%
\renewcommand{\arraystretch}{0.9}%
\begin{tabular}{l|cc|cc}
\toprule
\multirow{3}{*}{Method} & \multicolumn{2}{c|}{Orlando - Moderate} & \multicolumn{2}{c}{Hong Kong- Moderate} \\
\cmidrule{2-5}
 & Reward ↑ & Time ↓ & Reward ↑ & Time ↓ \\
 & Raw (Drop \%) & Raw (Rise \%) & Raw (Drop \%) & Raw (Rise \%) \\
\midrule
RELED (Full) & 782.34±2.81 & 505.21±21.13 & 1492.91±31.46 & 519.97±18.75 \\
 & (-) & (-) & (-) & (-) \\
 \midrule
RELED w/o Refn. & 627.51±90.20 & 553.68±24.57 & 1102.57±101.60 & 655.22±35.08 \\
 & (-19.79\%) & (+9.59\%) & (-26.15\%) & (+26.01\%) \\
\midrule
RELED-$\alpha_{0.2}$ & 513.13±115.46 & 623.31±31.06 & 1143.70±121.82 & 519.97±19.94 \\
 & (-34.41\%) & (+23.38\%) & (-23.39\%) & (+12.65\%) \\
\midrule
RELED-$\alpha_{0.5}$ & -75.51±359.01 & 1690.39±63.25 & 688.90±37.48 & 1708.52±55.19 \\
 & (-109.65\%) & (+234.59\%) & (-53.86\%) & (+228.58\%) \\
\midrule
RELED-Logit & 443.02±63.78 & 834.79±49.16 & 1060.71±70.87 & 770.90±41.58 \\
 & (-43.37\%) & (+65.24\%) & (-28.95\%) & (+48.26\%) \\
\midrule
IPPO & 481.58±135.40 & 682.86±39.22 & 865.21±35.76 & 1241.49±81.22 \\
 & (-38.44\%) & (+35.16\%) & (-42.05\%) & (+138.76\%) \\
\bottomrule
\end{tabular}
}%
\begin{tablenotes}
\item[$*$] Reported as mean ± SD over runs. The ↑ indicates higher value is better, while the ↓ indicates lower value is better.
\end{tablenotes}
\end{table}

\section{Conclusion}
\label{Sec: Conclusion}
We propose RELED, a scalable MARL framework that effectively integrates LLM-driven expert demonstrations with autonomous agent exploration. The SED module leverages theoretical non-stationarity bounds to guide the generation and refinement of high-quality expert trajectories, improving cumulative rewards and training stability. Furthermore, the HPO module adaptively balances learning from both expert- and agent-generated experiences, accelerating policy convergence and enhancing generalization for each agent. Experimental results show that RELED achieves better sample and time efficiency across scenarios, maintains performance advantages with increasing agent numbers in practical traffic management applications. As a potential future direction, we are looking forward to extending our
method to improve the performance of various applications such as robot control~\cite{zhang2025robust,zhang2025state,duan2025rethinking,wang2024agrnav,lin2025hasfl,wang2024he} and autonomous driving~\cite{lin2022channel,wang2020stmarl,lin2022tracking,fang2024ic3m}.

\bibliographystyle{IEEEtran}
\bibliography{reference}

\end{document}